\documentclass{article}

\PassOptionsToPackage{numbers, compress}{natbib}
\usepackage[preprint]{neurips_2026}

\usepackage[utf8]{inputenc} % allow utf-8 input
\usepackage[T1]{fontenc}    % use 8-bit T1 fonts
\usepackage{hyperref}       % hyperlinks
\usepackage{url}            % simple URL typesetting
\usepackage{booktabs}       % professional-quality tables
\usepackage{amsfonts}       % blackboard math symbols
\usepackage{nicefrac}       % compact symbols for 1/2, etc.
\usepackage{microtype}      % microtypography
\usepackage{xcolor}         % colors

\usepackage{hyperref}
\usepackage{graphicx}
\usepackage{float}
\usepackage{url}
\usepackage{amsthm}
\usepackage{amsmath}
\usepackage{caption}
\usepackage{subcaption}
\usepackage{comment}
\usepackage{booktabs}
\usepackage{siunitx}
\usepackage{multirow}
\usepackage{dsfont}
\usepackage{wrapfig}

\usepackage{algorithm}
\usepackage{algpseudocode}

\newcommand{\drift}{b}
\newcommand{\driftsigma}{b_{\sigma}}
\newcommand{\Jac}{J}
\newcommand{\RevJac}{\mathcal{A}}
\newcommand{\E}{\mathbb{E}}
\newcommand{\ip}[2]{\langle #1, #2 \rangle}
\newcommand{\binfield}{\tilde{x}}
\newcommand{\kmin}{k_{\mathrm{min}}}

\title{How Out-of-Equilibrium Phase Transitions can Seed Pattern Formation in Trained Diffusion Models}

\author{%
  Luca Ambrogioni \\
  Radboud University, \\ Donders Institute for Brain, Cognition, and Behaviour \\
}

\begin{document}
\maketitle

%\paragraph{Notation.} Throughout the paper, $t$ denotes the diffusion-time parameter in variance-preserving (VP) models, while $\sigma$ denotes the noise-scale parameter in variance-exploding or EDM-style models. We write $\drift(x,t)$ for the reverse drift in the VP setting, $\driftsigma(x,\sigma)$ for the reverse drift in the EDM setting, and $\Jac(x,t)=\partial \drift(x,t)/\partial x$ for the Jacobian of the reverse drift. We use $\mathbb{E}[\cdot]$ for probabilistic expectations and reserve angle brackets either for inner products, written explicitly as $\ip{\cdot}{\cdot}$, or for spatial averages when the averaging operation is clear from context. To avoid collision with the score notation $s_\theta$, the sign-binarized field is denoted by $\binfield=\operatorname{sign}(x)$ whenever it is introduced.

\begin{abstract}
Diffusion models generate structure by progressively transforming noise into data, yet the mechanisms underlying this transition remain poorly understood. In this work, we show that pattern formation in trained diffusion models can be explained as an out-of-equilibrium phase transition driven by instabilities in the denoising dynamics. We develop a theoretical framework linking data symmetries and architectural constraints, such as locality and translation equivariance, to the emergence of collective spatial modes. In this view, structure arises when low-frequency modes become unstable, triggering a rapid growth of spatial correlations that organizes noise into coherent patterns. We validate this theory through a combination of analytical models and experiments. In a controlled patch-based model, we observe a sharp softening of low-frequency modes in the theoretically predicted critical regime. Similar signatures are found in trained convolutional diffusion models on Fashion-MNIST and in large-scale ImageNet models, where pattern formation coincides with a pronounced weakening of spatial modes. Finally, intervention experiments show that applying guidance precisely at this critical stage improves class alignment compared to applying it at random times, demonstrating that this regime is not only descriptive but functionally important.

\end{abstract}

\section{Introduction}

Generative diffusion models have recently emerged as one of the most successful frameworks for high-fidelity image and video synthesis \citep{sohl2015deep,ho2020denoising,song2021score,karras2022elucidating}. Because these models gradually organize structure from disordered noise, they exhibit behavior reminiscent of physical systems. In statistical physics, the abrupt emergence of macroscopic order is often explained by \emph{spontaneous symmetry breaking phase transitions}, also known as critical phase transitions \citep{landau1937phase,kadanoff1966scaling,wilson1971rg,wilsonfisher1972epsilon,stanley1971phase}. A canonical example is magnetic ordering in a metal: although the underlying microscopic laws governing the atoms are rotationally symmetric, a magnet below its critical temperature spontaneously 'selects' a specific magnetic orientation through the amplification of microscopic fluctuations. This analogy raises a natural question: can symmetry-breaking mechanisms similarly explain pattern formation in generative diffusion models? A key indication comes from \citep{kamb2025creativity}, where it was shown that denoising dynamics based solely on locality and translational invariance can already produce complex patterns that closely match those generated by convolutional networks. This observation is particularly compelling, as symmetry and locality are central ingredients in the theory of critical phase transitions. In the last few years, several works show that reverse diffusion does not proceed as a smooth interpolation from noise to data, but rather evolves through distinct dynamical regimes. Several studies interpret the generative process in terms of phase-transition–like behavior, in which the noise-symmetric state becomes unstable and trajectories concentrate around modes of the data distribution \citep{biroli2024dynamical,raya2024symmetry,sakamoto2024geometry,esteban2026fisher,sclocchi2025phase,ambrogioni2025statistical,sclocchi2025probing}. Complementary approaches introduce the notion of \emph{critical windows}: short intervals of diffusion time during which specific features of the final sample are determined \citep{li2024criticalwindows,li2025blink}. Despite these advances, a precise connection to the theory of critical phase transitions remains lacking. Existing analyses typically focus either on low-dimensional pitchfork bifurcations \citep{raya2024symmetry,biroli2024dynamical} or on highly structured hierarchical models \citep{sclocchi2025phase,sclocchi2025probing}. Moreover, they are largely restricted to analytically tractable score functions, leaving open the question of how these insights extend to trained neural networks. Another important line of works focuses on the sampling of physical systems and computational complexity \citep{montanariwu2023posterior, elalaoui2022sampling, ghio2024sampling}, not on machine learning and image generation applications. 

\begin{wrapfigure}{r}{0.4\textwidth}
  \centering
  \includegraphics[width=0.38\textwidth]{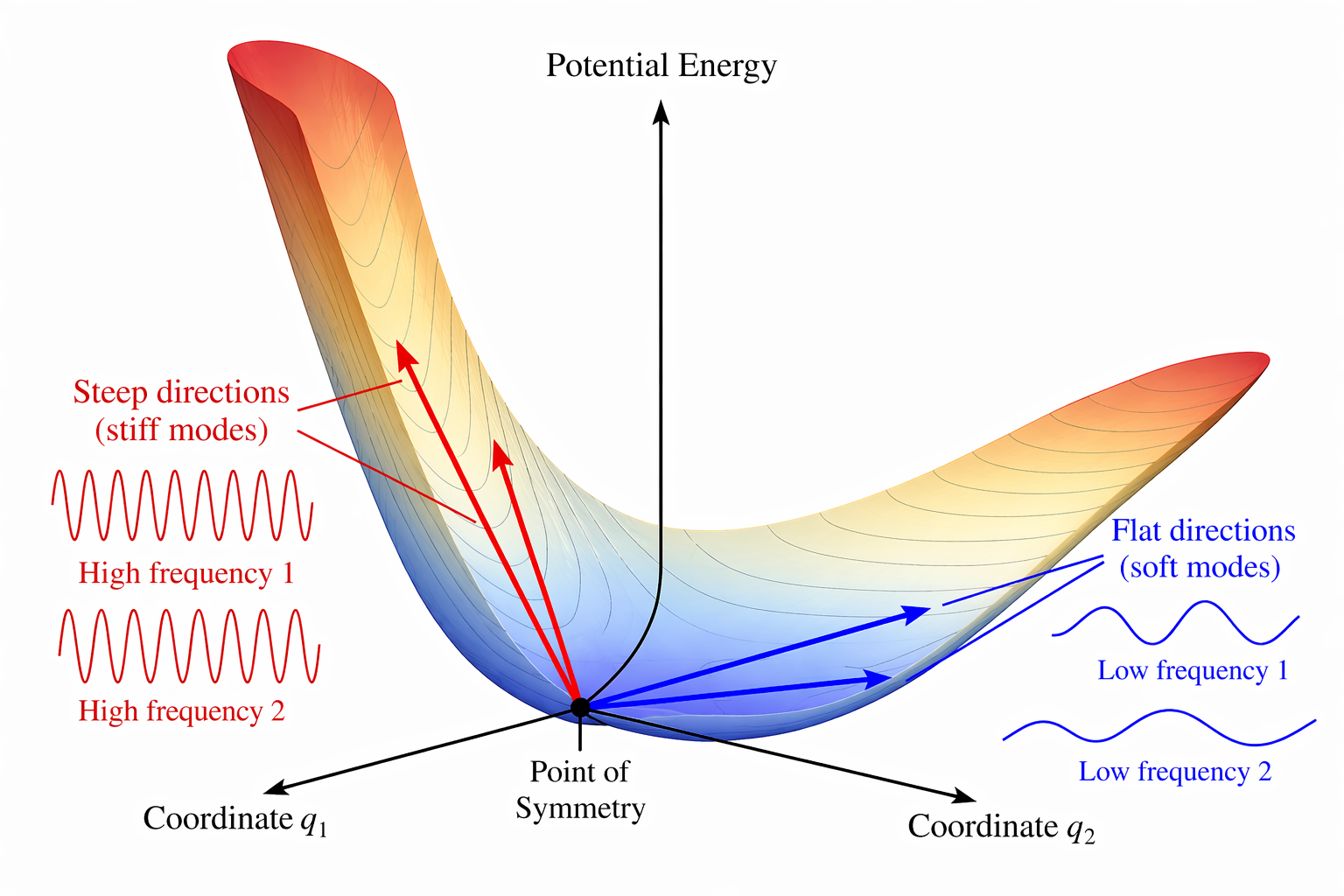}
  \caption{Visualization of mode softening. Low frequency modes lose stability at the critical points, leading to long range pattern formation.}
  \label{fig: mode softening}
\end{wrapfigure}

The key insight of this work is that the structure of these transitions arises from the interplay between symmetries in the data and architectural constraints of the network, which impose locality, sparsity, and invariance constraints. Building on this perspective, we analyze how pattern formation is triggered by sudden instabilities in low-frequency modes of the dynamics—referred to in physics as \emph{soft modes}. This \textbf{mode softening} phenomenon is a key prediction of our theoretical framework (visualized in Fig.~\ref{fig: mode softening}), as it is a reliable fingerprint of critical transitions in physical systems that can be directly studied in trained neural networks. We show that, in translationally invariant settings, these modes correspond to sinusoidal Fourier components. Their destabilization leads to the rapid emergence of structure, which is subsequently amplified by the nonlinear reverse diffusion dynamics. We describe this phenomenon as an out-of-equilibrium phase transition, as it is closely analogous to similar phenomena studied in cosmology and material science \citep{kibble1976topology}.

\begin{figure}[t]
\centering
\includegraphics[width=\linewidth]{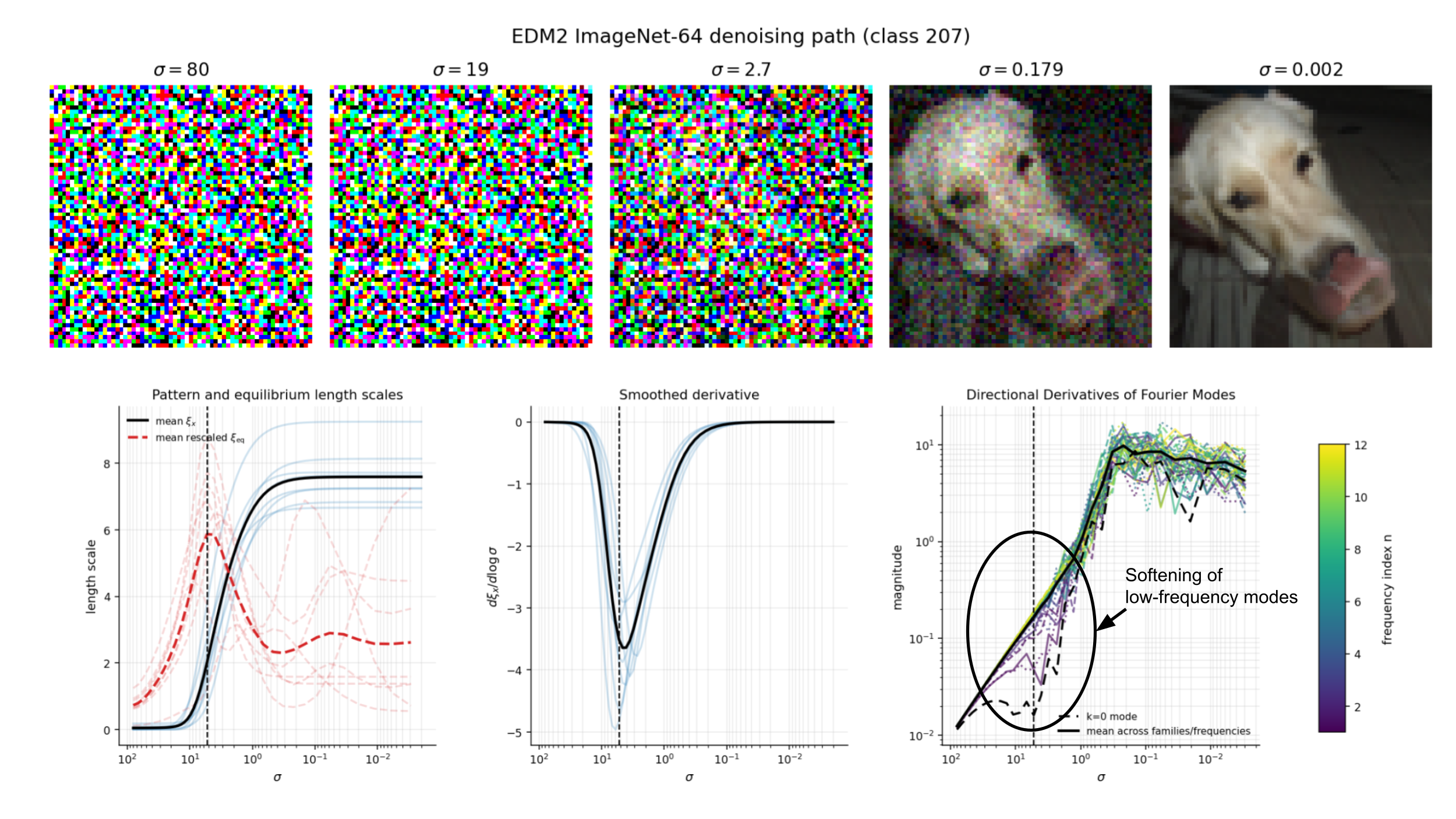}
\caption{
Quantitative analysis of pattern formation dynamics in a trained EDM2 model \citep{karras2022elucidating, karras2024analyzing}. Top: Example generative denoised trajectory.
Left: spatial correlation length $\xi_x(\sigma)$ along several reverse trajectories (transparent blue) and their median (black). The dashed red curves show the equilibrium correlation length estimate $\xi_{\rm eq}(\sigma)$ obtained from the normalized low-frequency dispersion of the reverse-drift Jacobian. 
Middle: derivative $d\xi_x/d\log \sigma$ highlighting the rapid growth of spatial coherence near the estimated critical scale. 
Right: magnitude of median directional derivatives (computed over $8$ trajectories) along a range of Fourier modes (colored) together with their median (black) and zero-th mode (dashed) showing the softening of spatial modes. Since EDM2 is variance exploding, all modes become arbitrarily low as $\sigma$ increases. However, the low frequency modes are strongly suppressed compared to high frequencies at the onset of pattern formation. 
}
\label{fig:edm2_results}
\end{figure}

%\paragraph{Contributions.} Our main contributions are:

%\begin{itemize}

%\item \textbf{Criticality as a mechanism for structure formation in diffusion models.}
%We show that the reverse diffusion process can pass through a \emph{critical dynamical regime} where spatial correlations rapidly grow and coherent structures emerge from noise. 

%\item \textbf{Architectural constraints transform memorization into collective generative dynamics.}
%We demonstrate that while unconstrained score models exhibit memorization-driven bifurcations tied to individual data points, architectural constraints such as \emph{locality and translation equivariance} convert these instabilities into instabilities in collective spatial modes (soft modes) capable of generating novel configurations.

%\item{\textbf{Guidance susceptibility.}} In ImageNet experiments, we show that pulses of classifier-free guidance \citep{ho2022classifierfree} delivered at the estimated "critical time" are more effective at aligning the generation to the class than random pulses, suggesting that the system is highly susceptible to external perturbations.

%\end{itemize}

\section{Background on symmetries}
We consider data points \( x = (x_i)_{i \in \Lambda} \) defined on a discrete spatial lattice \( \Lambda \) (e.g., pixels on an image grid). In the following, we will be particularly interested in the limit of large lattice size \( |\Lambda| \to \infty \), where collective and spatial effects become well-defined. A \emph{symmetry group} \( G \) is a set of transformations acting on configurations such that the distribution is invariant:
$
p_t(gx) = p_t(x), \quad \forall g \in G.
$
It is useful to distinguish two types of transformations. \emph{Global symmetries} act uniformly on the entire configuration (e.g., \( x \mapsto -x \)), while \emph{spatial symmetries} act on the lattice structure, such as translations \( x_i \mapsto x_{i+a} \). In many machine learning models, especially convolutional architectures, these spatial transformations are built into the model via equivariance. A point \( x^\ast \) is called \emph{symmetric} if it is invariant under all transformations:
$
g x^\ast = x^\ast, \quad \forall g \in G.
$
Such points correspond to maximally unstructured configurations. For example, in a dataset with reflection symmetry \( x \mapsto -x \), the configuration \( x = 0 \) is symmetric, while nonzero patterns are mapped into each other. In diffusion models under symmetries, symmetric points act as fixed-points of the dynamics and we are concerned with changes in their stability. While what we shall discus can be applied to any symmetry group, in all our analysis we will consider the simple symmetry $Z_2$:  \(x \mapsto -x\), which flips the value of binary pixels. 

\section{Background on bifurcations in generative diffusion models}
Diffusion models generate samples by learning to reverse a gradual
corruption of the data distribution with Gaussian noise.
Let $p_{\mathrm{data}}(x_0)$ denote the data distribution.
In the variance-preserving (VP) formulation \citep{song2021score}, the
forward process follows
$
dx_t =
-\tfrac12 \beta(t)x_t\,dt
+
\sqrt{\beta(t)}\,dW_t ,
$
whose solution can be written
$
x_t = \alpha(t)x_0 + \sigma(t)\epsilon,
\qquad
\epsilon \sim \mathcal N(0,I),
$
with
$
\alpha(t)=\exp\!\left(-\tfrac12\int_0^t \beta(s)\,ds\right),
\qquad
\sigma^2(t)=1-\alpha(t)^2 .
$
Given data $\{x_j\}_{j=1}^N$, a score network $s_\theta(x,t)$ is trained
with denoising score matching \citep{ho2020denoising}. The nonparametric optimum
corresponds to the empirical score
\begin{equation}
\nabla \log \hat{p}_t(x_t)
=
\sum_{j=1}^N
w_j(x_t)\,
\nabla_{x_t}\log p_t(x_t \mid x_0^j),
\end{equation}
where
$w_j(x_t)=p(x_0=x_0^j\mid x_t)\propto p_t(x_t\mid x_0^j)$
are normalized Bayesian weights. Once the score is learned, samples are generated by the reverse-time SDE
\begin{equation}
dx_t =
\Big[
-\tfrac12 \beta(t)x_t
-
\beta(t)s_\theta(x_t,t)
\Big]dt
+
\sqrt{\beta(t)}\,d\bar W_t .
\end{equation}
We denote the reverse drift in the VP setting by
$
\drift(x,t):=-\tfrac12 \beta(t)x-\beta(t)s_\theta(x,t),
$
so that the Jacobian of the reverse drift is $\Jac(x,t):=\partial \drift(x,t)/\partial x$. Since sampling proceeds from large to small $t$, it is convenient for stability analysis to introduce the increasing denoising-time parameter
$
\tau(t)=\int_t^T \beta(s)\,ds.
$
The deterministic drift per unit $\tau$ is then
$
\tilde b(x,t)=-\drift(x,t)/\beta(t)=\tfrac12 x+s_\theta(x,t),
$
and we denote its Jacobian by
$
\RevJac(x,t):=\partial \tilde b(x,t)/\partial x=-\Jac(x,t)/\beta(t).
$
Because $\beta(t)>0$, $\Jac$ and $\RevJac$ have the same zero crossings and normalized dispersions, with opposite signs. From a physics perspective, this is an annealed process since $t$ has both the role of dynamic time and effective temperature of the potential. As we shall see, this is what causes out-of-equilibrium phase transitions in the generative dynamics. We will now formalize the theory of symmetry breaking outlined in \citep{raya2024symmetry} and \citep{biroli2024dynamical, achilli2026theory, handke2026entropic}. For simplicity, consider a symmetric dataset consisting of two prototypes,
$
p_{\mathrm{data}}(x)
=
\frac{1}{2}\delta(x-\mu)
+
\frac{1}{2}\delta(x+\mu),
$
with $\mu \in \mathbb{R}^N$. After Gaussian noising with variance $\sigma^2(t)$, the forward marginal becomes a two-component Gaussian mixture whose log-density can be written as
\begin{equation}
\log \hat{p}_t(x)
=
-\frac{1}{2\sigma^2(t)}\|x\|^2
+
\log \cosh
\left(
\frac{\alpha(t)\ip{\mu}{x}}{\sigma^2(t)}
\right)
+
\mathrm{const}.
\end{equation}
The nonlinearity depends only on the global projection $\ip{\mu}{x}$, so the empirical geometry couples all coordinates through a single collective mode. Linearizing at the symmetric fixed point $x = 0$ yields the denoising-time stability operator
\begin{equation}
\RevJac(0,t)
=
\left(\frac{1}{2}-\frac{1}{\sigma^2(t)} \right) I
+
\frac{\alpha^2(t)}{\sigma^4(t)} \mu\mu^\top.
\end{equation}
Its spectrum consists of an $(N-1)$-fold degenerate stable eigenvalue and a single eigenvalue along the direction of $\mu$. The first instability occurs when this eigenvalue crosses zero, so the unstable manifold is one-dimensional. Projecting the deterministic drift component of the reverse dynamics onto this direction and defining
$
m = \frac{\ip{\mu}{x}}{\|\mu\|^2},
$
one obtains, to leading order, the denoising-time differential equation
\begin{equation}
\frac{d m}{d\tau}
=
- r(t)\, m
-
u(t)\, m^3 + O(m^5),
\qquad
r(t)
=
-\frac{1}{2}+\frac{1}{\sigma^2(t)}
-
\frac{\alpha^2(t)\|\mu\|^2}{\sigma^4(t)},
\qquad
u(t)=\frac{\alpha^4(t)\|\mu\|^6}{3\sigma^8(t)}.
\end{equation}
When $r(t)$ changes sign, a supercritical pitchfork bifurcation occurs with equilibria
$
m_\pm \simeq \pm\sqrt{-r(t)/u(t)}.
$

%Consider a symmetry group $G$ acting on $x_t$ such that the marginal density is invariant,
%\[
%p_t(gx_t)=p_t(x_t), \qquad \forall g\in G.
%\]
%A point $x^*$ is symmetric if $gx^*=x^*$ for all $g\in G$.
%In the two-pattern case, the symmetry reduces to a reflection $x \mapsto -x$, and the symmetric point $x=0$ is a fixed point of the reverse dynamics.

%In the following, we will often assume that the data respects this simple binary symmetry for the sake of simplicity. However, all methods can be easily extended to cases where the data has mode complex sets of values.

We now consider a dataset of $M$ equally weighted, equal-norm centered prototypes $\{\mu_m\}_{m=1}^M$, with $\sum_m \mu_m=0$. The point $x=0$ remains a symmetric fixed point, and the denoising-time stability operator takes the form
\begin{equation}
\RevJac(0,t)
=
\left(\frac{1}{2}-\frac{1}{\sigma^2(t)} \right) I
+
\frac{\alpha^2(t)}{\sigma^4(t)} C,
\qquad
C=\frac{1}{M}\sum_{m=1}^M \mu_m\mu_m^\top.
\end{equation}
Hence the stability is governed by the empirical second-moment matrix $C$.
Since $\RevJac(0,t)$ is an affine function of $C$, both share the same eigenvectors, and instabilities occur when
$
\lambda_k(t) = a(t) + c(t)\lambda_k(C)
$
cross zero. The unstable directions are therefore given by the leading eigenspaces of $C$. If the prototypes are approximately orthogonal, i.e.\ $\ip{\mu_m}{\mu_n} \approx 0$ for $m\neq n$, then each $\mu_m$ is an approximate eigenvector of $C$, so the unstable directions are approximately aligned with individual patterns. 

%In this regime, each instability can be interpreted as a partial selection of one prototype. In finite dimension, exact orthogonality is not possible, so this identification is only approximate. However, in high dimension, pairwise overlaps are small, and each branching induces only a weak perturbation of the remaining symmetric configuration, so the local description in terms of eigendirections remains approximately valid.

\section{From memorization to pattern formation through architectural constraints}
The goal of this section is to understand how architectural structure constrains the symmetry-breaking mechanisms that can appear in the learned dynamics. We begin by briefly recalling the learning setup. Let $\mathcal{F} = \{ s_\theta(\cdot,t) \mid \theta \in \Theta \}$ denote the class of score fields realizable by a given neural architecture. The parameters are learned by minimizing the denoising score-matching objective
\begin{equation}
\mathcal{L}(\theta) =
\mathbb{E}_{t,x_0,\epsilon}
\left[
\left\|
s_\theta(x_t,t) -
\nabla_{x_t}\log p_t(x_t \mid x_0)
\right\|^2
\right],
\end{equation}
where $x_t = \alpha(t)x_0 + \sigma(t)\epsilon$ is the forward diffusion process. In general, we can conceptualize the result of training as a non-linear projection of the empirical vector field onto the set of vector fields allowed by the model class:
$
    \mathcal{P}\left[ \nabla \log \hat p_t \right] = s_{\theta^*}(\cdot,t)
$
where $\theta^*$ depends implicitly on the global empirical score. The training dynamics of deep nets, as captured by the mapping $\mathcal{P}$, is extremely complex. However, the instability structure is fundamentally constrained by the architecture class, which can be used to infer several generic properties of this mapping.

If the function class $\mathcal{F}$ were sufficiently expressive, minimizing this loss would allow the network to recover the empirical score field associated with the dataset. However, in architectures with bounded receptive fields, each coordinate interacts only with a fixed-size neighborhood $\Omega$. The resulting Jacobian therefore contains nonzero entries only within a band determined by the receptive field. The number of independent couplings per site is $O(|\Omega|)$, where $|\Omega|$ is the number of sites in the receptive field. Consequently, the total number of free parameters in the linearized operator scales as
$
O(N |\Omega|),
$
rather than $O(N^2)$. Isolating a specific dense vector $v \in \mathbb{R}^N$, as an eigenvector of a matrix generically requires imposing $O(N)$ independent constraints on the operator entries. An even stronger restriction arises when the architecture is translation equivariant, as in convolutional score networks. In that case, in the large image limit, the Jacobian becomes a convolution operator. Such operators are diagonalized by spatial Fourier modes. Therefore
\begin{equation}
    \RevJac(0, t) = \sum_k \lambda(k,t) ~ e^{i k} e^{-i k^{\top}}.
\end{equation}
where the sum ranges all vectors of spatial frequencies $k$ allowed by the grid, forming a discrete Fourier basis over spatial modes. As standard in Fourier analysis, the complex exponential here is a convenient way to organize spatial sine and cosine waves. Note that $e^{i k}$ is a vector since $k$ is a vector.

From this diagonalization results, it follows that the instability structure of translationally invariant architectures (in the large image limit) is necessarily aligned with spatial frequency components, instead than with the individual data-points like in the empirical case. Writing the configuration as
$
x(r,t) = \sum_k q_k(t) e^{i k \cdot r},
$
the linearized dynamics decouples mode by mode,
\begin{equation}
    \frac{d q_k}{d\tau} = \lambda(k,t)\, q_k,
\end{equation}
where $\lambda(k,t)$ are the eigenvalues of the denoising-time stability operator $\RevJac$. The onset of pattern formation is thus controlled by the structure of this spectrum. As we shall see in the next section, for the inversion-symmetric local setting considered there, the spectrum is smooth at small $k$ and admits the expansion
$
\lambda(k,t) = -r(t) - \kappa(t) k^2 + O(k^4),
$
which defines an \emph{effective mass} $r(t)$ and \emph{stiffness} $\kappa(t)>0$. This jargon is borrowed from  physics, where $r$ plays the role of the mass of particles or particle-like perturbations (e.g. sound waves) since it determines the amount of energy needed to put them in motion. In this representation, unstable modes (soft modes) are precisely those for which $\lambda(k,t)$ approaches zero. The critical time $t_c$ is therefore defined by the condition $\lambda(0,t_c)=0$, or equivalently $r(t_c)=0$. The dynamics across this point is governed by a sign change of the mass term. For $r(t)>0$, all eigenvalues are negative and fluctuations decay. For $r(t)<0$, the modes acquire positive growth rates and grow exponentially. The instability is therefore not confined to a single mode but involves a finite band of low-frequency modes whose width increases with $|r(t)|$.  This band instability provides a direct mechanism for pattern formation. %Leading to the critical time, high wavelength modes start softening and spatial correlation grows. Immediately past criticality, only the longest wavelengths are unstable, leading to large-scale fluctuations. As the system evolves, more modes become unstable and nonlinear interactions select structured configurations. The selected pattern depends on the maximum of $\lambda(k,t)$: if it occurs at $k=0$, the system forms large domains, whereas a maximum at finite $k$ leads to patterns with a characteristic wavelength. 

%In generative diffusion, the system is out-of equilibrium, which means that the noise cannot fully propagate and cause an observable divergence of the correlation length of the fluctuations. Instead, out-of-equilibrium instabilities serve as seeds for subsequent pattern formation, driven by non-linearities acting upon the amplified noise fluctuations \citep{turing1952morphogenesis, swift1977hydrodynamic, kibble1976topology}. 

\section{Analysis of the patch score model}
\begin{figure}[t]
\centering
\includegraphics[width=\linewidth]{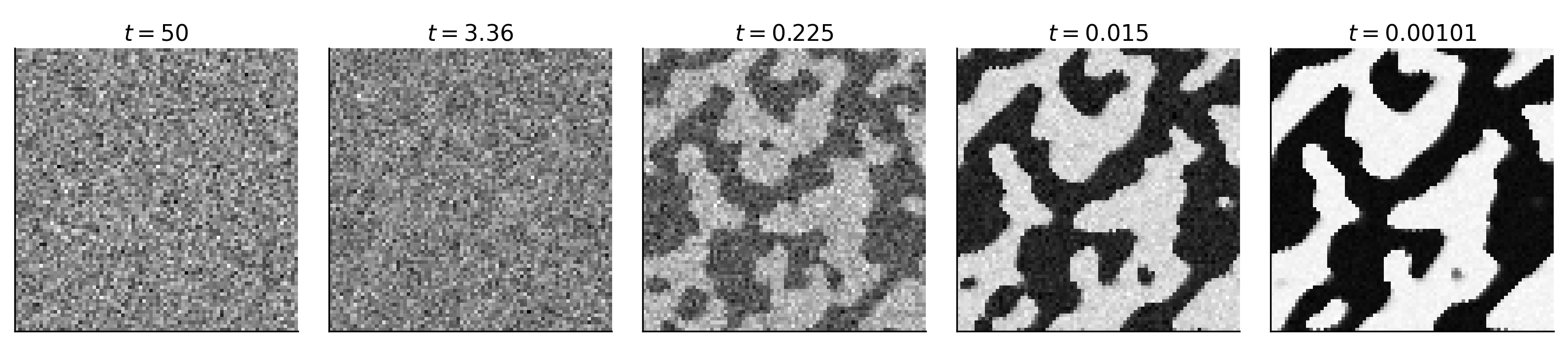}
\caption{
Five snapshots of the evolving configuration are shown at logarithmically spaced times along the reverse diffusion path. 
}
\label{fig:patch_results1}
\end{figure}
We now analyze a tractable analytic example illustrating how architectural locality converts global symmetry-breaking instabilities into spatially extended critical phenomena. We consider the minimal $\mathbb{Z}_2$-symmetric dataset $x_0 =  \pm \mathbf{1}$  with equal probability. The empirical score is
\begin{equation}
\nabla \log \hat{p}_t(x) = -\frac{x}{\sigma^2(t)} + \frac{\alpha(t)}{\sigma^2(t)} \E[x_0 \mid x]~,
\end{equation}
which is intrinsically nonlocal. Following \cite{kamb2025creativity}, we impose a locality constraint by restricting the score to
\begin{equation}
s_i(x,t) = f_t(x_{\Omega_i}), \qquad x_{\Omega_i} = \{x_{i+u} : u \in \Omega\}.
\end{equation}
where $f_t$ is an arbitrary function. The score-matching solution can then be computed analytically:
\begin{equation}
s_i(x,t) = \frac{1}{\sigma^2(t)}\left[\alpha(t)\tanh\!\left(\frac{\alpha(t)}{\sigma^2(t)} S_i(x)\right) - x_i\right],
\end{equation}
with $S_i(x) = \sum_{u\in\Omega} x_{i+u}$. We can now study the dynamics of this very simple 'learned' score model by analyzing the instability around the symmetric state $x^* = 0$.

A Taylor expansion around the symmetric state yields
\begin{equation}
s_i(x,t) = -\frac{x_i}{\sigma^2(t)} + \frac{\alpha^2(t)}{\sigma^4(t)} \sum_{u \in \Omega} x_{i+u}
- \frac{\alpha^4(t)}{3\sigma^8(t)} \left(\sum_{u \in \Omega} x_{i+u}\right)^3 + \cdots.
\label{eq:patch_expansion}
\end{equation}

To relate this expression to the soft-mode picture of Sec.~5, it is essential to consider the large-system limit. Around the symmetric point, the denoising-time stability operator is given by the convolution operator
\begin{equation}
    \RevJac_{ij}(0, t) = \left( \frac{1}{2} - \frac{1}{\sigma^2(t)}\right) \delta_{ij} + \frac{\alpha^2(t)}{\sigma^4(t)} \mathds{I}((j-i) \in \Omega)
\end{equation}
where $\mathds{I}$ is equal to one when the condition is respected and zero otherwise. In the thermodynamic limit with translationally invariant patches, $\RevJac(0,t)$ is diagonalized by Fourier modes. Writing the patch as a hyper-cubic neighborhood
$
\Omega = \{u \in \mathbb{Z}^d : |u_\mu|\le K,\ \mu=1,\dots,d\}~,
$
the eigenvectors are plane waves $e^{i k \cdot i}$ and the corresponding eigenvalues are
\begin{equation}
\lambda(k,t)
=
\left(\frac{1}{2} - \frac{1}{\sigma^2(t)}\right)
+
\frac{\alpha^2(t)}{\sigma^4(t)}
\sum_{u\in\Omega} e^{i k\cdot u}.
\label{eq:patch_lambda_general}
\end{equation}
The first term is the spatially uniform contribution from the Ornstein--Uhlenbeck drift and Gaussian part of the score, while the second term is the local convolution kernel induced by the architectural constraint. For the hypercubic patch, the Fourier symbol factorizes and the full analytic spectrum is
\begin{equation}
{
\lambda(k,t)
 = 
\left(\frac{1}{2} - \frac{1}{\sigma^2(t)}\right)
+
\frac{\alpha^2(t)}{\sigma^4(t)}
\prod_{\mu=1}^d
\frac{\sin\!\big((K+\tfrac12)k_\mu\big)}{\sin(k_\mu/2)}
}
\label{eq:patch_lambda_exact}
\end{equation}
with $k$ in the Brillouin zone. In particular, the most unstable mode is the uniform mode $k=0$, with the $\lambda(0,t)$ eigenvalue being given by
$
- \lambda(0,t) = r(t) =
-\frac{1}{2} + \frac{1}{\sigma^2(t)}
-
\frac{\alpha^2(t)}{\sigma^4(t)}(2K+1)^d.
\label{eq:patch_lambda_zero}
$
At the mean-field level, where we do not consider the effect of stochastic fluctuations around the symmetric point, this gives us the following expression for the critical noise level:
\begin{equation}
   \lambda(0,t_c) = 0 \Longrightarrow \left(-\frac{1}{2} + \frac{1}{\sigma^2(t_c)}\right)
=
\frac{\alpha^2(t_c)}{\sigma^4(t_c)}(2K+1)^d 
\end{equation}
To investigate soft non-zero frequency modes, we expand \eqref{eq:patch_lambda_exact} for small wave-vector, which gives the low-frequency dispersion of the stability operator
\begin{equation}
{
\lambda(k,t)
=
- r(t)
-
\kappa(t) |k|^2
+ O(|k|^4)
}.
\label{eq:patch_lambda_smallq}
\end{equation}
This shows that locality converts the mean-field bifurcation into a dispersive instability with a finite stiffness
$
\kappa(t)
=
\frac{\alpha^2(t)}{\sigma^4(t)}
\frac{K(K+1)}{6}(2K+1)^d~.
$
The instability therefore remains maximal at $k=0$, but extends over a continuum of nearby long-wavelength modes in the large-system limit, producing the spatially extended critical regime characteristic of local score architectures.

\subsection{Coarse graining, continuum dynamics and renormalization}
To fully relate this expression to the soft-mode picture of Sec.~5, it is useful to consider the large-system (continuum) limit. On a finite lattice, instabilities correspond to isolated eigenvalues of the linearized operator crossing zero. In contrast, in the large image limit, translation equivariance implies a dispersion relation $\lambda(k,t)$ that is smooth in $k$.  So instability occurs through the simultaneous softening of a band of low-frequency modes. This structure underlies the emergence of a growing correlation length and cannot be fully captured at the purely discrete level. For configurations varying slowly on the scale of $\Omega$, we expand
$
\sum_{u \in \Omega} x_{i+u} \approx |\Omega|\,x_i + c_\Omega \nabla^2 x_i,
\qquad c_\Omega=\frac{K(K+1)}{6}|\Omega| \quad \text{(hypercubic patch)},
$
which gives the denoising-time drift
$
s(x,t) + \frac{1}{2} x \approx -r(t)\,x + \kappa(t)\nabla^2 x - u(t)x^3,
$
with
$
u(t) = \frac{\alpha^4(t)}{3\sigma^8(t)}(2 K + 1)^{3d}.
$
In the continuous limit, the deterministic denoising dynamics per unit $\tau$ is $\partial_\tau x = \tfrac12 x + s(x,t)$ and therefore takes the form
\begin{equation}
\partial_\tau x = -r(t)\,x + \kappa(t)\nabla^2 x - u(t)x^3.
\end{equation}
This is a deterministic coarse-grained evolution equation that describes the dynamics of the system at low-frequencies, where we can neglect the effect of the original grid (e.g. the discrete pixels in an image).  The derivation above yields a deterministic dynamics in which the onset of the instability is controlled by the coefficient $r(t)$. In the stochastic reverse process, however, fluctuations modify this prediction. This is a profound topic that touches on the theory of renormalization and we are only going to consider it briefly at a superficial level. To make the effect of fluctuations explicit, we decompose the field into a smooth component and fluctuations,
$
x = \bar{x} + \delta x,
$
and expand the cubic term:
$
x^3 = \bar{x}^3 + 3 \bar{x}^2 \delta x + 3 \bar{x} (\delta x)^2 + (\delta x)^3.
$
Averaging over fluctuations at fixed $\bar{x}$ and using symmetry ($\langle \delta x \rangle = 0$) gives
$
\langle x^3 \rangle \approx \bar{x}^3 + 3 \bar{x} \langle \delta x^2 \rangle.
$
Substituting into the coarse–grained dynamics yields
\begin{equation}
\partial_\tau \bar{x} = -r(t)\bar{x} + \kappa(t)\nabla^2 \bar{x} - u(t)\bar{x}^3 - 3u(t)\langle \delta x^2 \rangle \bar{x}.
\end{equation}

Thus fluctuations re-normalize the linear coefficient, defining an effective mass term
\begin{equation}
r_{\mathrm{eff}}(t) = r(t) + 3u(t)\langle \delta x^2 \rangle.
\end{equation}
The variance $\langle \delta x^2 \rangle$ can be approximately estimated from the linearized dynamics. This leads to a growth of $\langle \delta x^2 \rangle$ near the instability, and therefore to a significant shift in the effective mass. Since $\langle \delta x^2 \rangle > 0$, the instability condition is shifted from $r(t)=0$ to $r_{\mathrm{eff}}(t)=0$, implying a shift to smaller $t$ (later along the reverse denoising trajectory) than the mean-field prediction. 

\section{Experiments on patch score model} \label{sec: patch score experiments 1}
We consider a two–dimensional patch model on a periodic $80\times 80$ lattice. 
The generative model consists of $5\times5$ binary patches drawn from a discrete dictionary. 
The dictionary contains eight random patterns sampled from $\{\pm1\}^{5\times5}$ together with their sign-flipped copies to enforce a global $\mathbb{Z}_2$ symmetry. In addition, two global patterns consisting of all $+1$ and all $-1$ patches are included. The prior probability mass is concentrated on the global patterns ($p=0.9$ total), while the eight random patterns share the remaining mass ($p=0.1$ total). We simulate the reverse dynamics of the variance–preserving diffusion process with constant noise rate $\beta=1$. 
The reverse SDE is integrated numerically on a logarithmic time grid from $T=50$ to $t=10^{-3}$ using $2000$ integration steps. 
The score function $s(x,t)$ is computed analytically from the full patch posterior without linearization. Fig.~\ref{fig:patch_results1} shows the patch dictionary together with a representative reverse denoising trajectory. 
At early times ($t\approx T$) the system is dominated by Gaussian noise. As time decreases, local patch evidence accumulates and coherent spatial domains begin to form. Eventually the system locks into large-scale binary regions corresponding to the dominant global patterns. 

\begin{figure}[t]
\centering
\includegraphics[width=\linewidth]{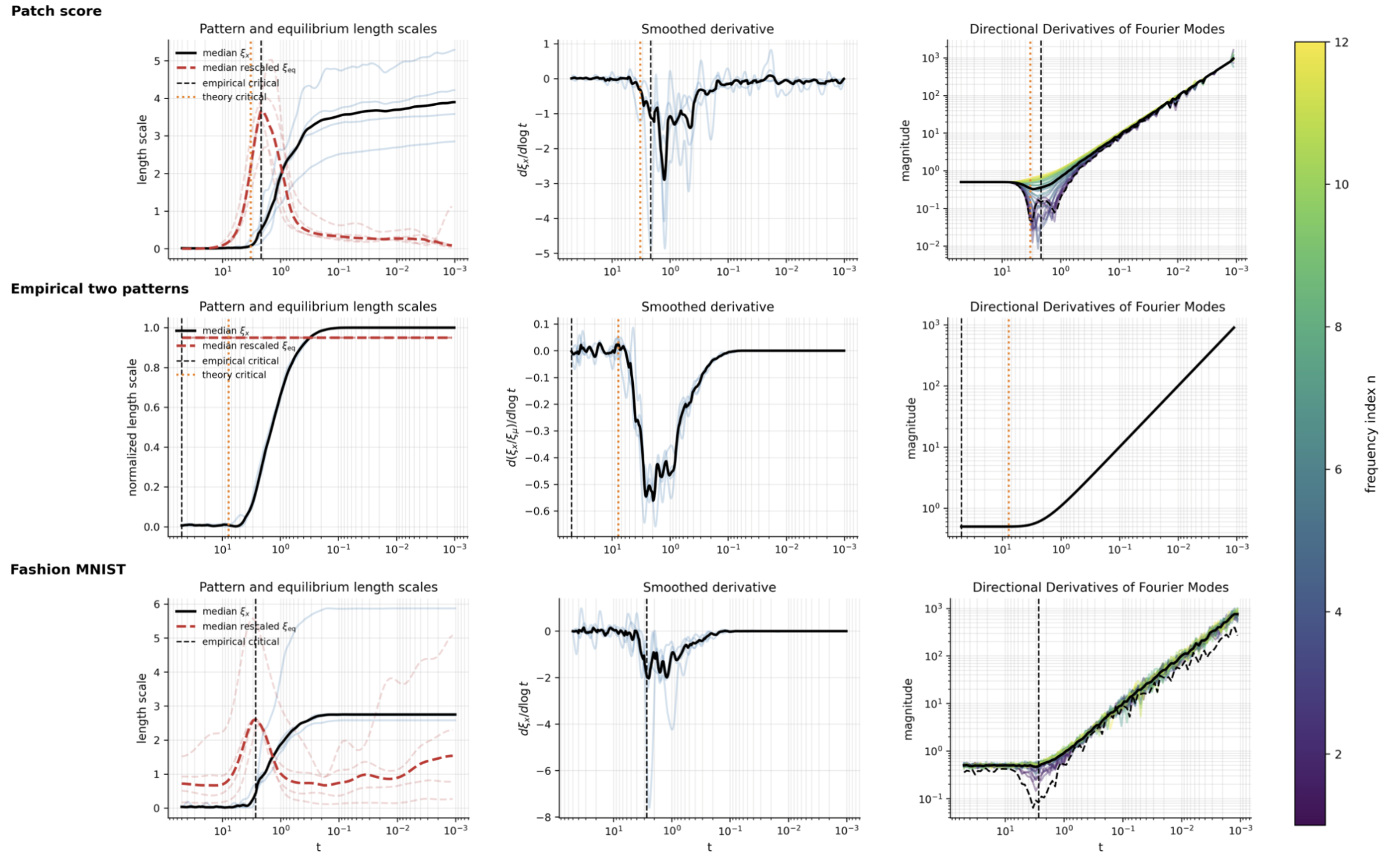}
\caption{
Quantitative analysis of pattern formation dynamics in patch score model (top), two patterns empirical score (middle), trained Fashion MNIST ConvNet (bottom)
Left: spatial correlation length $\xi_x(t)$ along several reverse trajectories (transparent blue) and their median (black). The dashed red curves show the equilibrium correlation length estimate $\xi_{\rm eq}(t)$ obtained from the normalized low-frequency dispersion of the reverse-drift Jacobian. 
Middle: derivative $d\xi_x/d\log t$ highlighting the rapid growth of spatial coherence near the critical time. 
Right: magnitude of median directional derivatives (computed over $8$ trajectories) along a range of Fourier modes (colored) together with their mean (black) and the zero-th frequency mode (dashed), showing the softening of spatial modes near the transition.
}
\label{fig:patch_results2}
\end{figure}
To quantify the emergence of spatial structure we measure the correlation length of the binarized field
$
\binfield=\mathrm{sign}(x)
$
along the reverse trajectory. 
The spatial correlation length $\xi_x(t)$ is estimated from the radially averaged autocorrelation function of $\binfield$. 
To obtain stable estimates of the temporal derivative we first smooth $\xi_x(t)$ with a Gaussian kernel before computing $d\xi_x/d\log t$. Fig.~\ref{fig:patch_results2} (left) shows $\xi_x(t)$ for several independent reverse trajectories (transparent curves) together with their median (black curve). 
The correlation length remains close to zero at large noise levels and increases rapidly once spatial structure begins to form. 
The middle panel shows the derivative $d\xi_x/d\log t$, which exhibits a pronounced peak near the onset of pattern formation. 
The vertical dashed line indicates the approximate mean-field critical time
$
t_c^{\mathrm{approx}} = \log(26),
$
obtained by neglecting the $\tfrac12$ Ornstein--Uhlenbeck drift contribution in the exact criticality condition. For convenience and consistency with the existing plots, we retain this approximate reference value; the exact mean-field expression is given in Appendix~\ref{app:gl-critical-time}, and the difference does not affect the qualitative comparison.

\paragraph{Linearized dynamics and equilibrium correlation length.}
To understand the origin of this transition we analyze the linearized reverse drift along the denoising trajectory. 
Directional derivatives of the drift are computed along Fourier modes
$
v_k(x,y)=\cos(kx),\qquad \cos(ky),\qquad \cos(k(x + y)).
$
These directional derivatives are Fourier-direction Rayleigh quotients of the Jacobian; at the translation-invariant symmetric branch they coincide with the corresponding Fourier eigenvalues. Since $\Jac=-\beta\RevJac$, their zero crossings and normalized dispersion are the same as those of the stability operator used in the theory. The right panel of Fig.~\ref{fig:patch_results2} shows the magnitude of these directional derivatives for several spatial modes. As the reverse process approaches the critical region the low-frequency modes soften progressively, producing a pronounced minimum near $t_c^{\mathrm{approx}}$. 
This softening signals the emergence of long-range spatial fluctuations. Using these measurements we estimate an approximate equilibrium correlation length $\xi_{\rm eq}(t)$ associated with fluctuations around the instantaneous denoising branch. 
To obtain a stable estimate we consider the normalized low-frequency dispersion of the Jacobian,
\begin{equation}
\frac{\lambda_n(t)}{\lambda_0(t)}
\approx
1 + \xi_{\rm eq}(t)^2 (n \kmin)^2 ,
\end{equation}
where $\lambda_n(t)$ denotes the directional derivative along the $n$-th Fourier shell and $\kmin=2\pi/L$. This normalization removes the rapidly increasing overall stiffness of the dynamics at small $t$ and isolates the spatial curvature of the spectrum. 
Fitting this relation across the lowest Fourier shells provides an estimate of $\xi_{\rm eq}(t)$. Note that this quantity is not a direct observable correlation due to the out-of-equilibrium nature of the dynamics, it is instead a proxy to quantify the softening of the modes at criticality. It should be therefore simply be interpreted as a quantification of the soft mode instability, not as a true thermodynamic quantity.

The resulting curves are shown as dashed red lines in Fig.~\ref{fig:patch_results2} (top). 
Individual trajectory estimates are shown in transparency while the median estimate is plotted in dashed red. The inferred equilibrium correlation length grows prior to the critical point, peaks close to the minimum of the soft modes, and subsequently decreases again as the dynamics locks into a stable pattern. This behavior explains why the observed pattern correlation length begins increasing before the critical time: the underlying spatial modes start softening earlier, amplifying long-wavelength fluctuations before the system fully transitions into a patterned state. On the other hand, in Fig.~\ref{fig:patch_results2} (middle) we see the same quantities plotted for an empirical score with two patterns $-\mu$ and $\mu$ with $\mu(x,y) = \text{sign}\left(\cos\left(\frac{2 \pi}{48}\left(x + 2y \right)\right)\right)$. In this case, the inferred correlation length is roughly constant and there is not visible softening of low frequency modes, with the critical times corresponding to a one-dimensional bifurcation.

\begin{figure}[t]
\centering
\includegraphics[width=\linewidth]{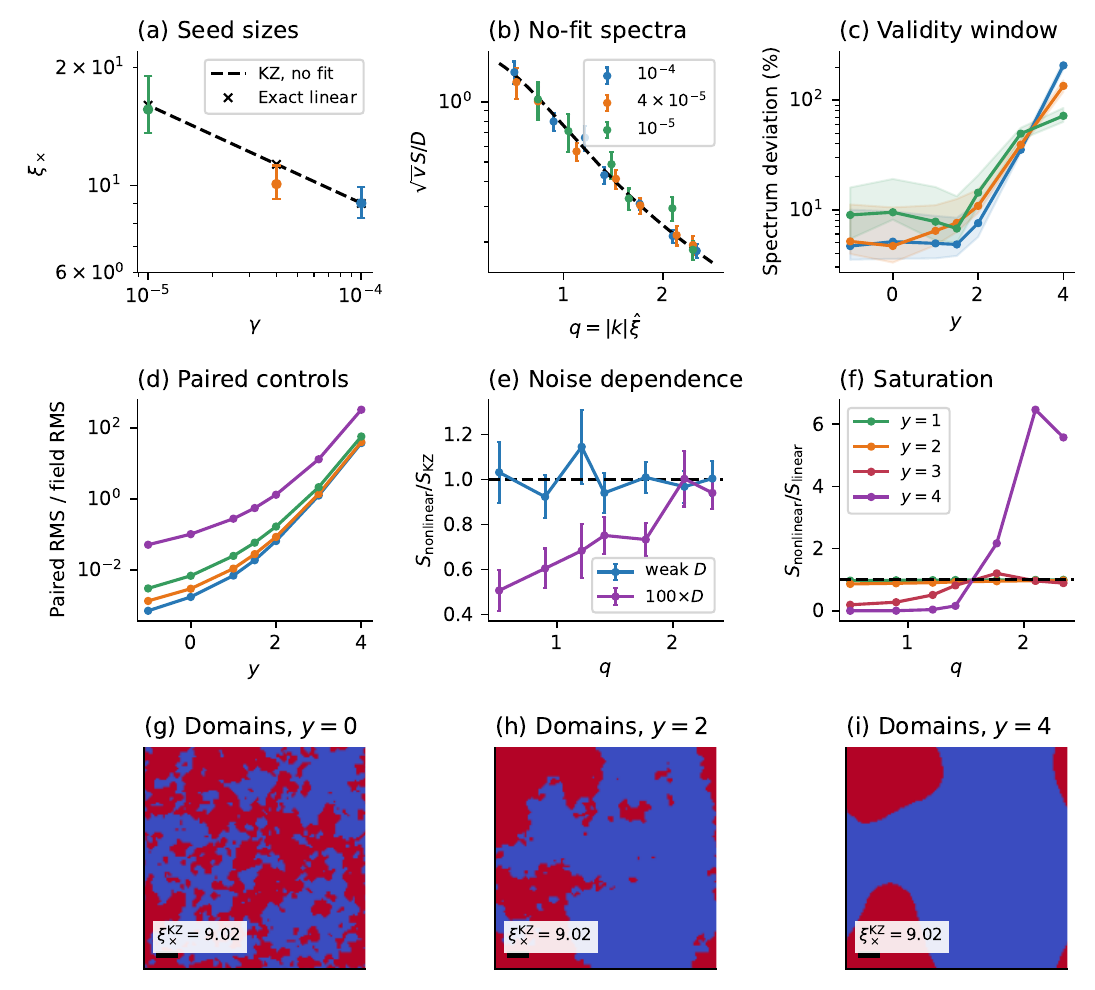}
\caption{\textbf{Quantitative tests of domain seeding near the symmetry-breaking point.} (a) Predicted KZ crossing length (dashed), exact intrinsic lattice linearization (crosses), and nonlinear spectral estimates (points; $95\%$ trajectory-bootstrap intervals). (b) Nonlinear crossing spectra at three rates and the predicted scaling function, without fitted parameters; bars show $1.96$ ensemble standard errors. Rate colors apply to (a--d). (c) Spectral deviations from the prediction through time, with bootstrap intervals. (d) Relative RMS difference between nonlinear and exact linear trajectories driven by the same noise; purple denotes $100$-fold stronger noise. (e) Crossing spectra at weak and stronger noise, with the same standard-error bars. (f) Nonlinear/paired-linear spectral ratios at later times, showing saturation and redistribution of power. (g--i) One weak-noise trajectory, smoothed by a one-site Gaussian filter before binarization for display; all spectra and lengths use unsmoothed fields. The fixed bars denote $\xi_\times^{\rm KZ}=9.02$, rather than a final domain diameter. Weak-noise ensembles contain $64$ trajectories per rate; the stronger-noise test contains $32$.}
\label{fig:kz-domain-prediction}
\end{figure}

% BEGIN KZ MAIN ADDITION
\begin{samepage}
\begingroup
\setlength{\parskip}{1pt}
\setlength{\abovedisplayskip}{3pt}
\setlength{\belowdisplayskip}{3pt}
\setlength{\abovedisplayshortskip}{2pt}
\setlength{\belowdisplayshortskip}{2pt}
\section{Predicting domain scales}
\label{sec:kz-domain-scales}
The preceding experiments relate the onset of spatial organization to the softening of low-frequency modes. The linearized description also provides a quantitative prediction for the correlation scale that seeds the emerging domains. To obtain this prediction, we return to the symmetric configuration $x=0$ and follow its stochastic fluctuations near the symmetry-breaking point. This gives a crossing length and spectrum whose validity can be tested against the full nonlinear dynamics.

We consider the two sign-symmetric ferromagnetic patches underlying Eq.~\eqref{eq:patch_expansion}, whose symmetric-state Jacobian has dispersion $\lambda(k,t)=-r(t)-\kappa(t)|k|^2+O(|k|^4)$, with $r(t_c)=0$. We evolve the full score according to $dx_i=[\tfrac12x_i+s_i(x,t)]d\tau+\sqrt{2D}\,dW_i$, evaluating its coefficients along $t=t_c-\gamma(\tau-\tau_c)$ for $\beta=1$. Small $D$ and $\gamma$ give a controlled weak-noise, slow-crossing regime; the native reverse SDE instead has $D=1/2$ and $\gamma=1$. Writing $v_0=r'(t_c)>0$, $v=\gamma v_0$ and $\kappa_c=\kappa(t_c)$, linearization in field amplitude and expansion near $t_c$ give
\begin{equation}
 \partial_\tau x_k=\left[v(\tau-\tau_c)-\kappa_c|k|^2\right]x_k+\eta_k.
 \label{eq:kz-main-protocol}
\end{equation}
On the stable side, relaxation takes time $1/|r|$, which increases as the crossing approaches. Equating it to the remaining driving time $|\tau-\tau_c|$ sets the Kibble--Zurek scales
\begin{equation}
 \hat\tau=v^{-1/2},\qquad \hat\xi=\sqrt{\kappa_c}\,v^{-1/4}.
 \label{eq:kz-main-scales}
\end{equation}
Thus a tenfold slower crossing predicts correlations approximately $1.78$ times larger, with $\hat\xi\propto\gamma^{-1/4}$ and an absolute scale fixed by the score coefficients. The linear equation further predicts the full power spectrum. Using orthonormal Fourier modes, define $S(k,\tau)=\mathbb E|x_k(\tau)|^2$, $q=|k|\hat\xi$ and $y=(\tau-\tau_c)/\hat\tau$. Integration from the stable side yields
\begin{equation}
 S(k,y)=\frac{D\sqrt\pi}{\sqrt v}\operatorname{erfcx}(q^2-y),\qquad
 \xi_\times^{\rm KZ}=\sqrt{\frac{2}{\sqrt\pi}}\,\hat\xi,
 \label{eq:kz-main-spectrum}
\end{equation}
where $\operatorname{erfcx}(z)=e^{z^2}\operatorname{erfc}(z)$ and $\xi_\times^2=-\partial_{|k|^2}\log S(k,0)|_{k=0}$. This continuous-field second-moment length quantifies the correlations available to seed domains. It differs from the sign-field estimator $\xi_x(t)$ used above and from individual domain diameters; testing the spectrum therefore checks more than a fitted length alone.

\paragraph{Numerical tests.}
Exact lattice linearized covariance propagation reproduces the crossing lengths within $0.13\%$, testing the continuum, near-critical reduction. To test its validity for the full score, we simulate $64$ nonlinear trajectories at each of $\gamma=10^{-4},4\times10^{-5},10^{-5}$, with $K=2$, $L=128$ and $D=2\times10^{-5}$. The predicted seed lengths are $9.02,11.34,16.04$ sites, compared with spectral estimates $9.02,10.09,15.64$ and trajectory-bootstrap uncertainty (Fig.~\ref{fig:kz-domain-prediction}a). Amplitude and shape length are fitted independently at each rate. The measured exponent is $-0.245$, with $95\%$ interval $[-0.344,-0.170]$, consistent with $-1/4$ at a precision insufficient to distinguish nearby exponents. Comparing the complete spectra with their independently predicted shape and amplitude provides a stronger test (b,c): mean absolute shell deviations remain approximately $5$--$10\%$ through $y=1.5$.

\paragraph{From critical correlations to nonlinear domains.}
Exact linear controls driven by the same noise isolate nonlinear corrections. At the crossing, their RMS difference from the full score is only $0.17$--$0.67\%$ of the field RMS (d). Increasing $D$ one hundredfold suppresses the crossing spectrum (e) and reduces the inferred seed length to $6.10$ sites ($95\%$ interval $[5.26,6.97]$), compared with the weak-noise prediction $9.02$. The symmetric-state Jacobian is therefore predictive while fluctuation corrections remain small. Even at weak noise, spectral deviations reach $35$--$50\%$ by $y=3$, as saturation changes the spectral shape beyond the independent linear-mode description (f). The snapshots (g--i) show the seeded correlations developing into larger domains; their bars mark the predicted crossing scale, rather than final diameters. Linearization predicts initial correlations and early growth, while later domain sizes require nonlinear evolution and coarsening. A curvature-driven wall model retaining the nonlinear score response predicts the squared-radius loss of driven circles within $0.1$--$2.1\%$. Network coarsening requires calibration, and native-generation validity remains unestablished (Supplement~\ref{supp:domain-walls}, Figs.~\ref{fig:wall-controlled}--\ref{fig:wall-driven}). Supplement~\ref{supp:kz} gives uncertainty estimates, numerical checks and the conditions $K\ll\hat\xi\ll L$ with weak fluctuations.
\endgroup
\end{samepage}

% END KZ MAIN ADDITION

\section{Experiments on real equivariant architectures} \label{sec: kibble-zurek experiments}
In supplementary section \ref{supp sec: convnet theory}, we show that critical transitions are present in arbitrary translationally equivariant and local architectures as far as the data respects some symmetries. Even when the symmetries are not exact, like in real images, the phenomenon of critical mode softening can be confirmed qualitatively on trained ConvNets. We trained a model on the dataset FashionMNIST \citep{xiao2017fashion}, the details are given in Supp.\ref{supp: networks training}. We binarized the data so as to preserve the same approximate Ising-like symmetry group. Fig.~\ref{fig:patch_results2} (bottom) shows that the results are strikingly similar to the theoretical model, with a clear peak of equilibrium correlation length at the estimated critical point, a general softening of low frequency modes in a critical region and a sudden onset of pattern formation starting at the estimated critical point. This qualitative agreement is significant. The patch model and the trained U-Net differ radically in their microscopic description: one is a simplified analytic system with explicit patch interactions, while the other is a high-dimensional non-linear neural network trained on real data with partial symmetries. Yet both systems exhibit the same dynamical signatures predicted by the theory, mode softening, abrupt spatial structure formation, and rapidly growing correlations. Supp. Fig~\ref{fig:fmnist_soft_results} shows that this qualitative behavior is approximately preserved even when the model is trained on non-binarized data (under the same specifications), showing that the phenomenology is robust to changes in the symmetry structure of the data. More strikingly, as shown in Fig.\ref{fig:edm2_results} given at the beginning of the paper, a similar phenomenology can be observed in very complex EDM2 models. In this EDM setting, the reverse dynamics is parameterized by the noise scale $\sigma$ rather than the VP time variable $t$, with $\sigma=\sigma(t)$ under the corresponding noise schedule. In this case, we again see a clear softening of the low frequency modes in a critical region. The inferred correlation length has a clear peak and corresponds with the onset of pattern formation. Note that the initial slope of the median mode response is due to the fact that EDM2 uses a variance exploding model, meaning that the system does not equilibrate and instead it spreads out, leading to a linear softening of all modes at large noise scale. The EDM2 experiment is described in detail in supp.\ref{supp sec: edm2}.

\begin{figure}[t]
\centering
\includegraphics[width=\linewidth]{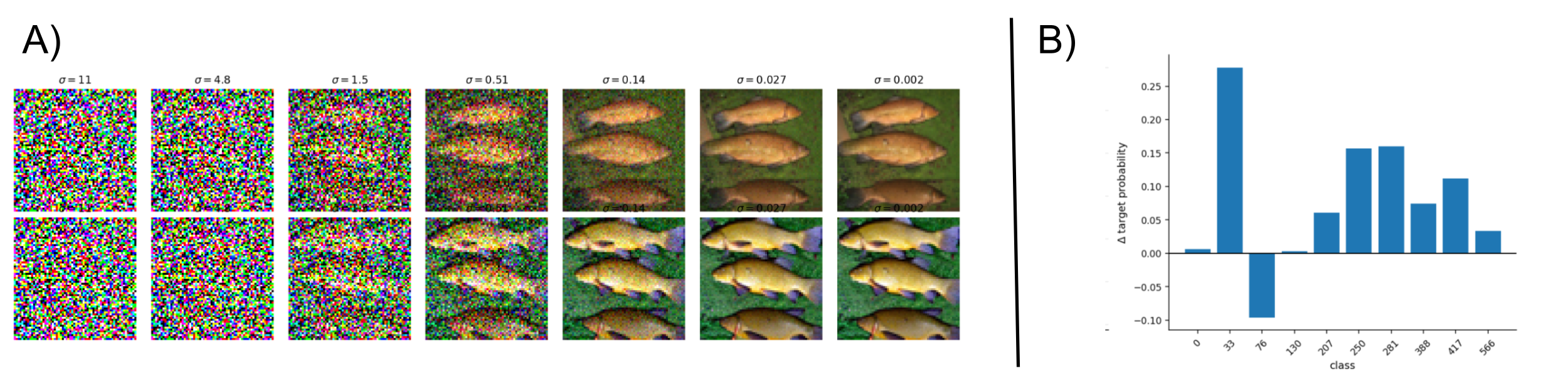}
\caption{
Results of pulse guidance experiment. A) Example image generated by the EDM2 model without critical guidance pulse (top) and with critical guidance pulse (bottom). B) Difference in class alignments (DINOV2 score) between critical pulse guidance and random pulse guidance for different classes. Critical pulses show higher alignment for almost all tested classes.
}
\label{fig:guidance_results}
\end{figure}

\textbf{Guidance pulse experiments.} To test whether the critical regime identified by our dynamical analysis is functionally relevant for generation, we performed a pulse-guidance intervention on a pretrained EDM2 ImageNet diffusion model. The key idea is to apply classifier-free guidance \citep{ho2022classifierfree} only within a short window of noise scales during the reverse trajectory, rather than throughout the entire sampling process. For each class, we first estimated the reverse-diffusion dynamics and extracted the equilibrium correlation length proxy $\xi_{\rm eq}(\sigma)$ from the normalized low-frequency dispersion of the Jacobian of the reverse drift. To reduce noise, the critical scale was not estimated separately for each class; instead, we aggregated the unguided trajectories across all classes and defined a single estimated global "critical" scale
$
\sigma_c = \arg\max_\sigma \xi_{\rm eq}(\sigma).
$
This provides an empirical estimate of the window in which long-wavelength modes are maximally softened. We then compared two intervention strategies. In the \emph{critical-pulse} condition, a strong pulse of classifier-free guidance was applied only within a narrow interval centered at $\sigma_c$. In the \emph{random-pulse} condition, an equally strong pulse with the same duration was applied at a random noise scale, sampled uniformly in $\sigma$. For each of 10 ImageNet classes we generated 40 samples per condition. Final samples were scored using a pretrained DINOv2 ImageNet classifier \citep{oquab2023dinov2}, and class alignment was quantified by the target-class logit and probability. Details are given in Supp.~\ref{supp sec: guidance experiments}. Fig.~\ref{fig:guidance_results} shows that, across classes, pulses delivered near the critical window systematically improved class alignment relative to pulses delivered at random times. This shows that the regime identified by the low-frequency softening analysis is not merely descriptive: it marks a stage of the reverse dynamics where perturbations have higher leverage over the emerging large-scale structure. 

\section{Conclusion and Limitations}
Our theory and results suggest that concepts from non-equilibrium statistical physics, instabilities, soft modes, and effective field theories provide a useful organizing framework for understanding trained diffusion models under architectural constraints such as translational equivariance. Overall, in diffusion models, the dynamics can always be locally described by a Jacobian and there is a clear unimodal potential in the high-noise limit, making instabilities essentially unavoidable except in trivial cases \citep{ambrogioni2025statistical, biroli2024dynamical}. However, the structure of these instabilities can differ significantly in models without a strong invariant spatial organization, like in latent diffusion models and stable diffusion. Further work is needed to extend this theoretical framework to more generic architectures and data structures, where the system is not approximately translationally invariant. It is also important to stress that the criticalities described in this paper are out-of-equilibrium instabilities, not equilibrium phase transitions. Since the studied systems are finite and the generative system does not fully thermalize, they give rise to finite cross-overs without a full divergence of thermodynamic potentials. This is formally very similar to Kibble-Zurek theory, where, like in generative diffusion, the control parameter (e.g. temperature) is linearly linked to the dynamic time \citep{kibble1976topology}.

%What matters is not whether a true phase transition exists in the finite system, but how these instabilities organize: in finite systems, they typically appear as crossovers rather than sharp critical points. This is entirely analogous to physics, where the thermodynamic limit is a conceptual tool since real systems are finite and out of equilibrium, yet still exhibit regimes influenced by universality classes. In real models, the value of this framework is therefore methodological and diagnostic: it provides concrete tools (spectra, correlations, mode structure) to identify when diffusion dynamics enter regimes that are well described by critical-like behavior, even if only approximately.

\section{Acknowledgments}
The author wishes to thank the participants of the Aspen Winter Conference Theoretical Physics for Artificial Intelligence (2026) for the insightful discussions that contributed to the ideas developed in this work. In particular, the author is grateful to Mason Kamb and Akhil Premkumar.

The author also acknowledges the role of GPT-5.4 in conceptual brainstorming, experimental implementation, and collaborative drafting of this manuscript. The system served as a high-fidelity intellectual partner, assisting in refining and evaluating the underlying physical theory, as well as in implementing the experimental code. However, the final synthesis, theoretical framing, and verification of all claims remain the sole responsibility of the author.

\bibliography{main}
\bibliographystyle{style.bst}

\appendix

\clearpage
\section*{Supplementary material: scope and reading guide}
Appendix~\ref{app:gl-critical-time} derives the coefficients and instability threshold of the two-uniform-patch model. Appendix~\ref{supp sec: convnet theory} states the structural assumptions behind the Fourier-mode analysis. The following methods sections describe the mixed patch dictionary, trained convolutional model, pretrained EDM2 analysis and guidance intervention. Supplement~\ref{supp:kz} then tests quantitative predictions for critical-window correlations and nonlinear walls. The trained-model diagnostics are qualitative comparisons; they do not establish that those models satisfy every assumption of the analytic theory.

\paragraph{Clocks and operator signs.}
The VP diffusion time $t$ decreases during generation. The denoising clock $\tau$ increases, with $d\tau=-\beta(t)dt$ for the native process. At fixed $t$, the deterministic denoising drift is $\widetilde b=\tfrac12x+s(x,t)$, while the backward-time drift is $\drift=-\beta(t)\widetilde b$. Consequently $\Jac=-\beta(t)\RevJac$: a stability sign must be interpreted with its clock. EDM sampling instead evolves with decreasing noise scale $\sigma$. The controlled schedules in Supplement~\ref{supp:kz} vary the coefficients independently of the denoising clock and are modified dynamics, not mere time reparameterizations.

\paragraph{Distinct measures of spatial scale.}
The symbol $\xi_x$ in the original experiments denotes an operational correlation statistic, whose definition depends on the field being analyzed. The quantities below are not interchangeable:
\begin{center}
\begin{tabular}{p{0.25\textwidth}p{0.67\textwidth}}
\toprule
Quantity & Definition and interpretation \\
\midrule
Sign-field length & First radial moment of the positive lobe of the sign-field autocorrelation. \\
Continuous-image length & Square root of the second radial moment of the positive correlation lobe of centered luminance. \\
$\xi_{\rm eq}$ proxy & Length-like coefficient of a normalized low-frequency Jacobian response; overlay rescaling removes absolute units and can affect ensemble peak selection. \\
Spectral seed length & Square root of $-\partial_{|k|^2}\log S(k)|_{k=0}$, predicted near the symmetry-breaking crossing. \\
Domain-network length & Inverse sign-boundary bond density, with its smoothing and lattice-direction conventions specified in Supplement~\ref{supp:domain-walls}. \\
\bottomrule
\end{tabular}
\end{center}
A Fourier-direction Rayleigh quotient coincides with an eigenvalue only when that direction is an eigenvector. Along heterogeneous generated fields, Fourier directions are probes of the Jacobian rather than an automatically diagonal basis. Plots of absolute response magnitudes can show softening but do not by themselves establish a signed eigenvalue crossing. Throughout the supplement, an exact model calculation is distinguished from a spatial or amplitude expansion, a fitted phenomenological closure and an empirical diagnostic. 

\clearpage

\section{Ginzburg--Landau coefficients and the mean-field threshold}\label{app:gl-critical-time}

We specialize to two equally weighted uniform patches on a scalar periodic lattice. The coarse-grained Ginzburg--Landau (GL) coefficients follow from a Taylor expansion in field amplitude and a second-order expansion of the symmetric patch convolution. They are bare mean-field coefficients, not fluctuation-renormalized parameters or coefficients inferred from a trained network. In the increasing denoising clock, they are
\begin{equation}
r(t)=-\frac{1}{2}+\frac{1}{\sigma^2(t)}-\frac{\alpha^2(t)}{\sigma^4(t)}|\Omega|,
\end{equation}

\begin{equation}
\kappa(t)=\frac{\alpha^2(t)}{\sigma^4(t)}\,c_\Omega,
\qquad
c_\Omega=\frac{K(K+1)}{6}|\Omega| \quad \text{for the hypercubic patch},
\end{equation}

\begin{equation}
u(t)=\frac{\alpha^4(t)}{3\sigma^8(t)}|\Omega|^3.
\end{equation}

For this specified patch prior, the GL parameters are determined by the forward diffusion statistics $(\alpha(t),\sigma(t))$ and the receptive-field geometry $\Omega$. At the mean-field level, instability occurs when the quadratic coefficient changes sign,
\begin{equation}
r(t_c)=0.
\end{equation}

Using the expression above gives the exact mean-field criticality condition
\begin{equation}
\begin{aligned}
\left(-\frac{1}{2}+\frac{1}{\sigma^2(t_c)}\right)
&=
\frac{\alpha^2(t_c)}{\sigma^4(t_c)}|\Omega|,\\
\alpha^4(t_c)+2|\Omega|\alpha^2(t_c)-1&=0.
\end{aligned}
\end{equation}
The physical root is
\begin{equation}
\alpha^2(t_c)=\sqrt{|\Omega|^2+1}-|\Omega|.
\end{equation}
Since
$\alpha^2(t)=\exp\!\left(-\int_0^t\beta(s)\,ds\right)$,
this gives
\begin{equation}
\begin{aligned}
\int_0^{t_c}\beta(s)\,ds
&=
\log\!\left(
|\Omega|+\sqrt{|\Omega|^2+1}
\right),\\
\beta(t)=1:\qquad
t_c
&=
\log\!\left(
|\Omega|+\sqrt{|\Omega|^2+1}
\right).
\end{aligned}
\end{equation}

The simpler expression used as a reference in the experimental plots is recovered by neglecting the $\tfrac12$ Ornstein--Uhlenbeck drift contribution in the criticality condition. In that approximation,
\begin{equation}
\begin{aligned}
\sigma^2(t_c)&\simeq\alpha^2(t_c)|\Omega|,
\qquad
\alpha^2(t_c)\simeq\frac{1}{1+|\Omega|},\\
t_c^{\mathrm{approx}}&=\log(1+|\Omega|)\qquad (\beta(t)=1).
\end{aligned}
\end{equation}
For the $5\times5$ patch, $|\Omega|=25$, giving $t_c\simeq3.91$ from the exact condition and $t_c^{\mathrm{approx}}=\log(26)\simeq3.26$. The original plots retain the approximate reference; quantitative predictions in Supplement~\ref{supp:kz} use the exact threshold. These two values must not be interchanged in timing tests. Neither is automatically the exact threshold of the mixed patch dictionary described below.

\section{Soft modes under equivariance and locality assumptions}
\label{supp sec: convnet theory}
This section gives a conditional description of the linear instability of a symmetric branch. It does not assert that every trained architecture has such a branch or undergoes the same bifurcation. Consider a real scalar field on the periodic lattice $\Lambda=(\mathbb Z/L\mathbb Z)^d$ and, at fixed diffusion time, a vector field $s_t:\mathbb R^\Lambda\to\mathbb R^\Lambda$. Assume differentiability through cubic order near $x=0$, dependence on a finite neighborhood $\Omega_R$, equivariance under every lattice translation, and global sign symmetry $s_t(-x)=-s_t(x)$. For a real dispersion, additionally assume that the linear kernel of the denoising drift is inversion symmetric. Global sign symmetry is distinct from independent sign flips at individual sites.

\subsection{Linearization and the soft sector}
The frozen denoising drift $\widetilde b_t(x)=\tfrac12x+s_t(x)$ vanishes at $x=0$ under these assumptions. Its Jacobian is a finite-range convolution,
\[
 (H(t)y)_i=\sum_{u\in\Omega_R}K_t(u)y_{i+u},\qquad
 \lambda(k,t)=\sum_{u\in\Omega_R}K_t(u)e^{ik\cdot u}.
\]
Translation equivariance at this homogeneous branch makes Fourier modes eigenvectors. Inversion symmetry $K_t(u)=K_t(-u)$ makes $\lambda$ real and even. Without it, Fourier diagonalization still holds, but linear stability is determined by $\operatorname{Re}\lambda$. For multichannel fields the Fourier symbol is matrix-valued, so stability requires its band eigenvalues rather than a single scalar Fourier response.

If a stable branch loses stability as the control parameter changes, its threshold satisfies
\[
 \max_k\lambda(k,t_c)=0,\qquad
 \mathcal M=\operatorname*{arg\,max}_k\lambda(k,t_c).
\]
This condition identifies marginal modes; it does not itself prove that a threshold is crossed. For an isolated maximizing wavevector $k_\star$ that is a stationary point of the continuous Fourier symbol, the local expansion is
\[
 \lambda(k,t)=-r(t)-\tfrac12(k-k_\star)^\top A(t)(k-k_\star)
 +O(|k-k_\star|^3),
\]
with $r(t_c)=0$ and $A(t)$ positive semidefinite. At an inversion-symmetric maximum at zero, odd spatial terms vanish and the remainder starts at fourth order. Evenness about zero does not generally eliminate cubic terms about a nonzero maximum. Degenerate maxima with vanishing quadratic curvature require higher-order spatial terms.

\subsection{Nonlinear amplitude description}
A slowly varying amplitude reduction additionally requires separation from noncritical modes, small amplitudes and spatial variation on scales exceeding the interaction radius. These conditions are not consequences of translation equivariance alone. For an isolated maximum at zero, the leading scalar equation has the form
\[
 \partial_\tau\phi=-r(t)\phi+\kappa_{\mu\nu}(t)\partial_\mu\partial_\nu\phi-u(t)\phi^3+\cdots.
\]
Symmetry permits the cubic term but does not determine its sign. Positive $u$ gives the usual supercritical saturation; other signs require higher nonlinearities. When the retained terms have variational dynamics with positive mobility, one may write the corresponding instantaneous Landau functional,
\[
 \mathcal F[\phi]=\int d^dz\,
 \left[\tfrac12r(t)\phi^2+
 \tfrac12\kappa_{\mu\nu}(t)\partial_\mu\phi\,\partial_\nu\phi+
 \tfrac14u(t)\phi^4+\cdots\right].
\]
Reciprocity of the linear kernel alone does not establish a variational representation of the full nonlinear dynamics, nor an equilibrium interpretation of this driven functional.

For a nonzero soft wavevector, reality requires including its conjugate partner:
\[
 x_i\simeq\psi(z)e^{ik_\star\cdot i}+\psi(z)^*e^{-ik_\star\cdot i}.
\]
The amplitude is generally complex. A real amplitude can describe a self-conjugate lattice mode, such as an appropriate zone-boundary mode; commensurability alone is insufficient. Several symmetry-related maxima require coupled amplitudes and their allowed nonlinear interactions. The linearization identifies candidate ordering channels; nonlinear coefficients determine their saturation and competition.

\subsection{Relation to the trained-model experiments}
The assumptions above describe an idealized local, reciprocal and symmetric field. The small U-Net uses zero padding, strided operations and GroupNorm. These operations do not ensure exact equivariance to every lattice shift or strict finite-radius dependence, and binarizing Fashion-MNIST does not enforce sign invariance of its distribution. The pretrained model is described with attention and class conditioning. Consequently their Fourier-direction measurements are empirical diagnostics of spatial response, rather than proofs of the scalar homogeneous reduction. This distinction leaves the qualitative comparison meaningful while restricting the quantitative claims to models and regimes where the assumptions are checked.

\section{Methods for the mixed patch model}

This section describes in detail the numerical experiments used to study the emergence of spatial structure in the reverse dynamics of the patch diffusion model. The analytic two-uniform-patch theory and the mixed dictionary used for these descriptive experiments are distinct models. 

\subsection{Patch dictionary}

The generative model is defined using a discrete dictionary of binary $5\times5$ patches. 
Each patch takes values in $\{-1,1\}^{5\times5}$. 

The dictionary contains:
\begin{itemize}
\item Four independently drawn patterns $p_k \in \{-1,1\}^{5\times5}$ in the implementation.
\item Their four sign-flipped counterparts $-p_k$, giving eight random dictionary entries with paired priors.
\item Two uniform patterns consisting of all $+1$ and all $-1$ entries.
\end{itemize}

The resulting dictionary therefore contains ten patterns.

The prior probability distribution over patches is defined as

\begin{equation}
\pi_k =
\begin{cases}
\frac{0.1}{8}, & k=1,\ldots,8 \\
\frac{0.9}{2}, & k=9,10 .
\end{cases}
\end{equation}

Thus the random patches collectively carry probability $0.1$ while the two uniform patterns together carry probability $0.9$.

The prior places most mass on uniform patches. This does not by itself determine the full nonlinear pattern-selection dynamics.

\subsection{Spatial domain}

The model is defined on a two–dimensional periodic lattice

\[
x \in \mathbb{R}^{L\times L}
\]

with size $L=80$ for the experiments in Sec.~\ref{sec: patch score experiments 1}. The experiments in Sec.~\ref{sec: kibble-zurek experiments} instead uses $L=48$; these configurations are distinguished below.

Periodic boundary conditions are used when extracting patch neighborhoods and computing spatial correlations.

\subsection{Forward diffusion process}

The forward diffusion process follows the variance-preserving stochastic differential equation

\begin{equation}
dx_t = -\frac12 x_t\,dt + dW_t .
\end{equation}

The corresponding marginal distribution satisfies

\begin{equation}
x_t = \alpha(t)x_0 + \sigma(t)\varepsilon,
\end{equation}

where

\begin{align}
\alpha(t) &= e^{-t/2} \\
\sigma(t) &= \sqrt{1-e^{-t}} .
\end{align}

\subsection{Exact patch score}

The central-pixel posterior score is computed analytically for the patch mixture. Its exactness is local to that specified mixture; assembling predictions from overlapping neighborhoods is not, without an additional integrability argument, an exact global image-distribution score. 
For each spatial location $(i,j)$ we consider the $5\times5$ neighborhood centered at that location.

Let $y_{ij}$ denote the patch extracted from the current state $x_t$. 

The posterior probability of patch $p_k$ is

\begin{equation}
w_k(i,j) \propto
\pi_k
\exp\left(
\frac{\alpha(t)}{\sigma(t)^2}
\ip{p_k}{y_{ij}}
\right) .
\end{equation}

After normalization,

\begin{equation}
\sum_k w_k(i,j)=1 .
\end{equation}

The posterior mean of the central pixel of the patch is

\begin{equation}
m_{ij} =
\sum_k w_k(i,j)\, p_k(0,0).
\end{equation}

The analytic score is then

\begin{equation}
s(x_t,t) =
\frac{\alpha(t)m - x_t}{\sigma(t)^2}.
\end{equation}

\subsection{Reverse diffusion dynamics}

The reverse SDE is

\begin{equation}
dx_t =
\left(
-\frac12 x_t - s(x_t,t)
\right) dt + d\bar W_t .
\end{equation}

Integration proceeds backward in time from

$T = 50$

to

$t_{\min}=10^{-3}$.

The experiments in Sec.~\ref{sec: patch score experiments 1} use logarithmic grid with $2000$ steps; the analysis in Sec.~\ref{sec: kibble-zurek experiments} instead uses $1200$ steps, with:

\begin{equation}
t_k = \exp\left(
\log T + \frac{k}{N}
(\log t_{\min}-\log T)
\right).
\end{equation}

The SDE is discretized using the Euler–Maruyama scheme

\begin{equation}
x_{k+1} =
x_k + b(x_k,t_k)\Delta t
+ \sqrt{|\Delta t|}\,\eta_k ,
\end{equation}

where

\[
\drift(x,t)= -\frac12 x - s(x,t).
\]

\subsection{Pattern correlation length}

Spatial structure is quantified using the correlation length of the binarized field

\begin{equation}
\binfield=\mathrm{sign}(x).
\end{equation}

This is an operational first-moment statistic of the sign field, not the spectral-curvature length tested in Supplement~\ref{supp:kz}.

The spatial autocorrelation function is computed via FFT:

\begin{equation}
C(r)=\left\langle\binfield_i\binfield_{i+\delta}\right\rangle_{i,\,|\delta|\in[r,r+1)}.
\end{equation}

A radial average is then computed to obtain $C(r)$ as a function of distance.

The implementation normalizes the radial correlation by $C(0)$ and truncates at its first nonpositive value, at radius $r_+$. It uses radial bins out to half the image width. The first positive-lobe moment is

\begin{equation}
\xi_x =
\frac{\sum_{r<r_+} r\,C(r)}{\sum_{r<r_+} C(r)}.
\end{equation}

\subsection{Temporal derivative}

To stabilize the estimate of

\[
\frac{d\xi_x}{d\log t}
\]

the correlation length trajectory is smoothed using a Gaussian kernel before differentiation.

\subsection{Linearized dynamics}

To analyze the stability of spatial modes we estimate directional derivatives of the reverse drift

\[
\Jac(x,t)=\frac{\partial \drift(x,t)}{\partial x}.
\]

In the implementation, perturbation directions are normalized to unit Euclidean norm before applying the stated finite-difference step. For a direction $v$ the directional derivative is approximated using symmetric finite differences

\begin{equation}
\lambda_v =
\frac{v^\top \left[
\drift(x+\varepsilon v,t)
-
\drift(x-\varepsilon v,t)
\right]}
{2\varepsilon\,v^\top v}.
\end{equation}

The finite difference step is chosen adaptively as

\[
\varepsilon = \max(5\times10^{-5}, 10^{-3}\sigma(t)^2).
\]

\subsection{Fourier modes}

Directional derivatives are computed along Fourier modes

\begin{align}
v_{n,x}(i,j) &= \cos(n \kmin i) \\
v_{n,y}(i,j) &= \cos(n \kmin j)
\end{align}

with

$\kmin = \frac{2\pi}{L}$.

Modes up to shell $n=6$ are evaluated.

\subsection{Jacobian dispersion-length proxy}

To estimate the equilibrium correlation length of fluctuations around the denoising branch we analyze the low-frequency dispersion of the Jacobian.

Let $\lambda_n(t)$ denote the directional derivative for shell $n$.

To remove the rapidly increasing stiffness of the dynamics at small $t$, the spectrum is normalized by the constant mode:

\begin{equation}
\tilde\lambda_n(t)
=
\frac{\lambda_n(t)}{\lambda_0(t)} .
\end{equation}

For small wavenumbers the dispersion is expected to follow

\begin{equation}
\tilde\lambda_n(t)
\approx
1 + \xi_{\mathrm{eq}}(t)^2 (n \kmin)^2 .
\end{equation}

The equilibrium correlation length is therefore estimated by fitting

\begin{equation}
\tilde\lambda_n(t)-1
=
\xi_{\mathrm{eq}}(t)^2 (n \kmin)^2
\end{equation}

over $n=1,2,3$ for the experiments in Sec.~\ref{sec: patch score experiments 1}. The implementation uses diagonal indices through $n=6$, as recorded below.

The fitted quantity is a dispersion-length proxy. Along a heterogeneous trajectory, the Fourier directions need not be eigenvectors, and the fit is not a thermodynamic equilibrium measurement. Signed responses are smoothed before fitting; nearly vanishing zero-mode responses and nonpositive fitted slopes are excluded, and finite estimates are interpolated and smoothed. Per-trajectory rescaling removes absolute-length information. Although it leaves each trajectory's peak position unchanged, it can change the relative weighting of curves and the peak of their ensemble summary. For a diagonal direction $\cos[n k_{\min}(i+j)]$, the physical wave-number magnitude is $\sqrt2 n k_{\min}$; for the experiments in Sec.~\ref{sec: patch score experiments 1}, the fit uses $n k_{\min}$ as an index. Its plotted proxy therefore must not be interpreted as an independently calibrated physical length.

\subsection{Trajectory statistics}

Individual trajectories are shown with transparency and their median in black. Dispersion-length curves are rescaled for visual overlays; their amplitudes cannot then be compared as independent length measurements.

\subsection{Critical time}

For the $5\times5$ patch and the constant schedule $\beta(t)=1$, the exact mean-field critical time derived in Appendix~\ref{app:gl-critical-time} and the approximate reference value used in the figures are
\begin{equation}
t_c \simeq 3.91,
\qquad
t_c^{\mathrm{approx}}=\log(26)\simeq3.26.
\end{equation}
For convenience and consistency with the existing figures, the vertical dashed line uses $t_c^{\mathrm{approx}}$, obtained by neglecting the $\tfrac12$ Ornstein--Uhlenbeck drift contribution. This is a uniform-patch reference, not an established exact threshold of the mixed dictionary. Its offset from the exact uniform-patch value matters for quantitative timing.

\subsection{Summary of parameters}

The following table records the parameter choices for the experiments in Sec.~\ref{sec: patch score experiments 1}; the settings are provided in the next subsection.

\begin{center}
\begin{tabular}{ll}
Grid size & $80\times80$ \\
Patch size & $5\times5$ \\
Random patches & $8$ \\
Uniform patches & $2$ \\
Reverse time range & $T=50$ to $10^{-3}$ \\
Integration steps & $2000$ \\
Trajectories & $4$ \\
Fourier shells & $n\le6$
\end{tabular}
\end{center}

These are the patch-method settings for the experiments in Sec.~\ref{sec: patch score experiments 1}; they do not specify the trained-model, guidance or controlled KZ experiments.

\section{Methods for the trained convolutional model} \label{supp: networks training}

This section describes in detail the experimental setup used to analyze pattern formation in trained diffusion models. The goal of these experiments is to reproduce, in a trained neural network, the dynamical phenomena observed in the analytic patch model. In particular, we measure the evolution of spatial correlations, the softening of Fourier modes of the reverse drift, and the effective equilibrium correlation length along the reverse diffusion trajectory.

\subsection{Dataset and preprocessing}

The experiments use the Fashion-MNIST dataset, which consists of grayscale images of clothing items with spatial resolution
$28 \times 28$.

To simplify the statistical structure of the data and emphasize spatial pattern formation, we use a binarized version of the dataset. Each pixel is converted to a binary variable using a deterministic threshold:
\[
x_{ij} =
\begin{cases}
+1 & \text{if } p_{ij} > 0.5, \\
-1 & \text{otherwise},
\end{cases}
\]
where $p_{ij}$ denotes the normalized grayscale pixel value.

After binarization the images satisfy
$x \in \{-1,1\}^{28\times28}$.

This preprocessing produces a binary field, but does not impose a sign-invariant image distribution. The comparison with a symmetric lattice model is therefore qualitative.

\subsection{Diffusion process}

We train the model using the variance-preserving (VP) diffusion formulation introduced in the main text. The forward diffusion process satisfies the stochastic differential equation
\[
dx_t = -\frac12 x_t \, dt + dW_t,
\]
corresponding to the constant noise schedule
$\beta(t) = 1$.

The solution of this process can be written in closed form as
\[
x_t = \alpha(t)x_0 + \sigma(t)\varepsilon,
\quad
\varepsilon \sim \mathcal{N}(0,I),
\]
with
\[
\alpha(t) = e^{-t/2},
\qquad 
\sigma^2(t) = 1 - e^{-t}.
\]

These functions determine the statistics of the noisy data distribution used during score-matching training.

The reverse-time diffusion process is given by
\[
dx_t =
\left(
-\frac12 x_t - s_\theta(x_t,t)
\right)dt + d\bar W_t,
\]
where $s_\theta(x,t)$ denotes the learned score function.

\subsection{Network architecture}
The model is a single-channel U-Net with base width $64$ and time-embedding dimension $128$. A sinusoidal embedding is processed by two linear layers with an intervening SiLU activation. Residual blocks apply GroupNorm, SiLU and a $3\times3$ convolution twice, add a projected time embedding after the first convolution, and use a $1\times1$ shortcut when channels change.
\begin{center}
\begin{tabular}{ll}
\toprule
Stage & Implementation \\
\midrule
Input & $3\times3$ convolution, $1\to64$ channels \\
Encoder & Residual blocks at $64$ and $128$ channels \\
Downsampling & $4\times4$ convolutions, stride $2$ \\
Bottleneck & Two residual blocks at $128$ channels \\
Decoder & Two transposed-convolution and skip-concatenation stages \\
Output & GroupNorm, SiLU, $3\times3$ convolution to one channel \\
\bottomrule
\end{tabular}
\end{center}
The network predicts noise $\varepsilon_\theta(x_t,t)$, converted to a score by
\[
 s_\theta(x_t,t)=-\varepsilon_\theta(x_t,t)/\sigma(t).
\]
The architecture supports spatial diagnostics but is not the exactly local, globally odd, periodic field assumed in Appendix~\ref{supp sec: convnet theory}.

\subsection{Training procedure}

The network is trained using denoising score matching with the standard noise prediction loss

\[
L(\theta) =
\mathbb{E}
\left[
\|
\varepsilon - \varepsilon_\theta(x_t,t)
\|^2
\right].
\]

Training parameters:

\begin{center}
\begin{tabular}{ll}
Dataset & Fashion-MNIST (binarized) \\
Epochs & 30 \\
Batch size & 256 \\
Optimizer & AdamW \\
Learning rate & $2\times10^{-4}$ \\
Gradient clipping & 1.0 \\
Time sampling & log-uniform
\end{tabular}
\end{center}

Log-uniform sampling over diffusion time ensures sufficient coverage of the small-noise regime, where the reverse dynamics becomes stiff.

\subsection{Reverse diffusion sampling}

Samples are generated by numerically integrating the reverse SDE on a logarithmic time grid

\[
t_k =
\exp\left(
\log T +
\frac{k}{N}
(\log t_{\min}-\log T)
\right),
\]

with the parameters for the experiments in Sec.~\ref{sec: patch score experiments 1}: %TODO: fix reference

\[
T = 50,
\quad
t_{\min}=10^{-3},
\quad
N=2000.
\]

The Euler–Maruyama scheme is used for integration.

\subsection{Spatial correlation analysis}

To analyze spatial pattern formation we measure the correlation length of the binarized configuration

\[
\binfield = \operatorname{sign}(x).
\]

This again corresponds to a first-moment estimator of the correlation length.

The spatial correlation function is computed as

\[
C(r;t) =
\langle \binfield_i \binfield_j \rangle_{|i-j|=r}.
\]

With $r_+$ the first nonpositive radial correlation in the implementation, the positive-lobe first moment is

\[
\xi_x(t) =
\frac{\sum_{r<r_+} r\,C(r)}{\sum_{r<r_+} C(r)}.
\]

To stabilize the derivative, the correlation-length trajectory is smoothed with a Gaussian kernel before computing

\[
\frac{d\xi_x}{d\log t}.
\]

\subsection{Linearized drift analysis}

To probe the stability of spatial modes we measure directional derivatives of the reverse drift

\[
\drift(x,t) =
-\frac12 x - s_\theta(x,t).
\]

In the implementation, perturbation directions are normalized to unit Euclidean norm before applying the stated finite-difference step. For a perturbation direction $v$ we estimate

\[
\lambda_v =
\frac{v^\top
\big(
\drift(x+\epsilon v,t) -
\drift(x-\epsilon v,t)
\big)}
{2\epsilon\,v^\top v}.
\]

This quantity is a Rayleigh quotient of the Jacobian of the reverse drift and coincides with an eigenvalue when $v$ is an eigenvector.

\subsection{Fourier modes}

Directional derivatives are computed along spatial Fourier modes

\[
v_{n,x}(i,j) = \cos(n \kmin i),
\quad
v_{n,y}(i,j) = \cos(n \kmin j),
\]

where

$\kmin = \frac{2\pi}{L}$.

\subsection{Jacobian dispersion-length proxy}

Following the analysis of the patch model we estimate an equilibrium correlation length from the normalized dispersion relation

\[
\frac{\lambda_n(t)}{\lambda_0(t)}
\approx
1 + \xi_{\rm eq}(t)^2 (n \kmin)^2.
\]

Fitting this relation over the lowest Fourier shells provides an estimate of $\xi_{\rm eq}(t)$.

The fitted quantity is a dispersion-length proxy. Along a heterogeneous trajectory, the Fourier directions need not be eigenvectors, and the fit is not a thermodynamic equilibrium measurement. In the experiments in Sec.~\ref{sec: kibble-zurek experiments}, signed responses are smoothed before fitting; nearly vanishing zero-mode responses and nonpositive fitted slopes are excluded, and finite estimates are interpolated and smoothed. Per-trajectory rescaling removes absolute-length information. Although it leaves each trajectory's peak position unchanged, it can change the relative weighting of curves and the peak of their ensemble summary. For a diagonal direction $\cos[n k_{\min}(i+j)]$, the physical wave-number magnitude is $\sqrt2 n k_{\min}$; the fit uses $n k_{\min}$ as an index. Its plotted proxy therefore must not be interpreted as an independently calibrated physical length.

\subsection{Trajectory statistics}

All observables are computed across several independent reverse trajectories. Plots show individual trajectories as transparent curves and their median as a solid curve. These descriptive curves show sample-to-sample variation; they are not confidence intervals for an instability threshold.

\subsection{Interpretation}

The convolutional architecture motivates spatial-response diagnostics, but its normalization, padding and striding do not satisfy all assumptions of Appendix~\ref{supp sec: convnet theory}. The measurements test qualitative spatial organization; they do not establish that the homogeneous theoretical operator exactly describes the trained network.

The reported trajectories show softening of long-wavelength responses, growing spatial correlations and the emergence of coherent structure.

These observations are consistent with the interpretation of diffusion generation as a nonequilibrium dynamical phase transition.

\subsection{Implementation scope}
The U-Net implementation uses zero-padded convolutions, strided downsampling, transposed-convolution upsampling and GroupNorm. The FFT analysis uses periodic correlations, which does not make the network boundary conditions periodic. The training configuration additionally specifies AdamW weight decay $10^{-4}$ and training times in $[10^{-3},50]$. Its final sampling call uses $1200$ steps. The finite-difference rule is $\epsilon=\max(5\times10^{-5},10^{-3}\sigma^2(t))$ and the proxy fit uses the first six diagonal directions. 

%These are verified implementation settings, not a reconstructed provenance record for every figure. The non-binarized result in the additional figures does not have a separately identified training configuration in the supplied notebook; its preprocessing and checkpoint should not be inferred from the binary experiment.

\section{Methods for the pretrained EDM2 analysis} \label{supp sec: edm2}

To test whether the theoretical predictions developed in this work extend to realistic large-scale diffusion models, we analyze the reverse diffusion dynamics of a pretrained pixel-space diffusion model trained on ImageNet. The critical line is estimated empirically as the peak of the inferred correlation length. In contrast to the analytic patch model and the small convolutional experiments described in the main text, this model operates on natural images and is trained on a large dataset using a modern diffusion architecture.

Our goal is to determine whether the same phenomenology observed in the analytic model and in small trained networks also appears in a state-of-the-art generative model. In particular, we investigate whether the reverse diffusion trajectory exhibits the signatures predicted by the statistical–physics theory:

\begin{itemize}
\item softening of low-frequency spatial modes,
\item growth of spatial correlation length during denoising,
\item emergence of a critical window where correlations are maximal.
\end{itemize}

\subsection{Model}

We use a pretrained pixel-space diffusion model from the EDM2 framework. The model is trained on the ImageNet dataset at a spatial resolution of $64 \times 64$ pixels. The model is described as a convolutional U-Net with attention and class conditioning.

The model predicts the denoised image $D_\theta(x,\sigma,y)$, where $x$ is the current noisy image, $\sigma$ is the noise level, and $y$ is the class label. The reverse-time dynamics of the sampler can be written as

\[
\frac{dx}{d\sigma} = \driftsigma(x,\sigma)
\]

with drift field

\[
\driftsigma(x,\sigma) =
\frac{x - D_\theta(x,\sigma,y)}{\sigma}.
\]

The variable $\sigma$ therefore plays the role of the control parameter along the generation trajectory, replacing the diffusion time $t$ used in the variance-preserving formulation discussed in the main text.

\subsection{Reverse diffusion trajectories}

Reverse diffusion trajectories are generated using the deterministic EDM sampler with Heun correction. The recorded positive-noise trajectory starts at

$\sigma_{\max} = 80$

down to

$\sigma_{\min} = 0.002$.

The noise schedule is parameterized by

\[
\sigma_i =
\left(
\sigma_{\max}^{1/\rho}
+
\frac{i}{N-1}
(\sigma_{\min}^{1/\rho} - \sigma_{\max}^{1/\rho})
\right)^{\rho}
\]

with $\rho = 7$. This schedule concentrates integration steps near the low-noise regime where the dynamics becomes stiff.

\subsection{Continuous spatial correlation length}

In the analytic patch model and the binarized Fashion-MNIST experiments, spatial correlations were measured using a sign-binarized field. For natural images this estimator is too coarse, as it discards most amplitude information and artificially introduces sharp domain boundaries.

Instead we compute the correlation length from a continuous scalar image field.

First, RGB images are converted to luminance

\[
\ell(x) =
0.2989 R + 0.5870 G + 0.1140 B.
\]

The spatial mean is subtracted to obtain a centered field

\[
\tilde{\ell}(x) =
\ell(x) - \langle \ell(x) \rangle.
\]

The connected spatial correlation function is then computed as

\[
C(r;\sigma)
=
\langle
\tilde{\ell}(x)\tilde{\ell}(x+r)
\rangle.
\]

Let $r_+$ be the first nonpositive radial correlation. The implementation uses the positive-lobe radial second moment

\[
\xi_x(\sigma)
=
\sqrt{
\frac{\sum_{r<r_+} r^2 C(r;\sigma)}
{\sum_{r<r_+} C(r;\sigma)}
}.
\]

The implementation truncates the normalized radial correlation at its first nonpositive value. Thus the sums above are over its positive lobe, using unweighted radial-bin means. This is a continuous-image radial moment, not an unrestricted thermodynamic second-moment length or the spectral curvature used in Supplement~\ref{supp:kz}.

\subsection{Linearized drift operator}

To probe the stability of spatial modes we analyze the Jacobian of the reverse drift field

\[
\driftsigma(x,\sigma) =
\frac{x - D_\theta(x,\sigma,y)}{\sigma}.
\]

In the implementation, perturbation directions are normalized to unit Euclidean norm before applying the stated finite-difference step. Directional derivatives of the Jacobian are computed using finite differences. For a perturbation direction $v$ we estimate

\[
\lambda_v(\sigma)
=
\frac{
v^\top
\left[
\driftsigma(x+\epsilon v,\sigma)
-
\driftsigma(x-\epsilon v,\sigma)
\right]
}
{2\epsilon\,v^\top v}.
\]

This quantity approximates the curvature of the drift operator along the direction $v$.

\subsection{Fourier-mode analysis}

To analyze spatial instabilities we evaluate directional derivatives along Fourier modes

\[
v_{n,x}(i,j) = \cos(n\kmin i),
\qquad
v_{n,y}(i,j) = \cos(n\kmin j)
\]

with

\[
\kmin = \frac{2\pi}{L}
\]

where $L$ is the image size.

Increasing $n$ probes increasing wave number and decreasing wavelength. The measured quantities are Fourier-direction Rayleigh quotients; they represent a dispersion only under an appropriate approximately diagonal description.

\subsection{Jacobian dispersion-length proxy}

Following the analysis developed for the analytic model, we estimate an equilibrium correlation length from the normalized dispersion relation

\[
\frac{\lambda_n(\sigma)}{\lambda_0(\sigma)}
\approx
1 +
\xi_{\mathrm{eq}}(\sigma)^2
(n\kmin)^2.
\]

This normalization removes the overall stiffness of the dynamics and isolates the curvature of the spatial spectrum near $k=0$.

The estimator is obtained by fitting the low-frequency modes of the dispersion relation.

\subsection{Empirical critical scale}

Unlike the analytic patch model, the trained neural network does not admit a closed-form expression for the coefficient $r(\sigma)$ of the effective Ginzburg–Landau theory. Consequently the critical point cannot be determined analytically.

Let $\Xi_{\rm proxy}$ denote the ensemble summary of the processed proxy curves. The empirical scale is its maximum:

\[
\sigma_c =
\arg\max_\sigma
\Xi_{\rm proxy}(\sigma),
\]

This operational peak identifies a candidate softened-response region. Equivalence to a zero of an effective mass requires additional assumptions about the operator, estimator and reference branch and is not established by the peak alone.

The fitted quantity is a dispersion-length proxy. Along a heterogeneous trajectory, the Fourier directions need not be eigenvectors, and the fit is not a thermodynamic equilibrium measurement. Signed responses are smoothed before fitting; nearly vanishing zero-mode responses and nonpositive fitted slopes are excluded, and finite estimates are interpolated and smoothed. Per-trajectory rescaling removes absolute-length information. Although it leaves each trajectory's peak position unchanged, it can change the relative weighting of curves and the peak of their ensemble summary. For a diagonal direction $\cos[n k_{\min}(i+j)]$, the physical wave-number magnitude is $\sqrt2 n k_{\min}$; the fit uses $n k_{\min}$ as an index. Its plotted proxy therefore must not be interpreted as an independently calibrated physical length.

\subsection{Trajectory statistics}

All observables are computed across multiple independent reverse diffusion trajectories.

Plots display:

\begin{itemize}
\item individual trajectories as transparent curves,
\item the ensemble mean of the length and its derivative as a solid curve.
\end{itemize}

\subsection{Interpretation}

The results show that the reverse diffusion dynamics of a large pretrained ImageNet model exhibit the same qualitative behavior predicted by the statistical–physics framework. As the noise level decreases, long-wavelength spatial modes soften and the spatial correlation length grows, indicating the emergence of coherent structure across the image.

The correlation length reaches a maximum within a narrow window of noise levels before decreasing again as the system enters a rigid low-noise regime. This behavior is consistent with the nonequilibrium critical dynamics described in the main text, where the reverse diffusion trajectory passes through a rounded critical regime during which spatial correlations are amplified before the system settles into a stable configuration.

\subsection{Analysis configuration}
We use the preset \texttt{edm2-img64-s-fid}, class index $207$, eight trajectories with seeds $1000$--$1007$, and $64$ sampler steps. It observes Jacobian responses every second recorded state and evaluates spatial indices through $n=12$, applying the same spatial perturbation to all RGB channels. The finite-difference scale is $\epsilon=\max(2\times10^{-5},2\times10^{-4}\sigma^2)$. The proxy uses the constant and first three diagonal directions, with Gaussian index width $1.5$ before fitting and after interpolation. The continuous-image length is smoothed with width three before differentiation in $\log\sigma$. Display overlays rescale each finite proxy curve to $0.95$ times that trajectory's maximum image length. These transformations are diagnostic and visual conventions, not absolute-length predictions.

\section{Methods for the guidance-pulse intervention} \label{supp sec: guidance experiments}

This section describes the guidance pulse experiments performed on a pretrained EDM2 ImageNet diffusion model. The objective of these experiments is to test whether the critical regime identified by the dynamical analysis of the reverse process also corresponds to a window of enhanced controllability. Specifically, we ask whether a short pulse of classifier-free guidance has a stronger effect when applied near the empirically identified critical scale than when applied at a random point along the denoising trajectory.

\subsection{Model and sampling framework}

We use a pretrained pixel-space EDM2 diffusion model on ImageNet at resolution $64\times64$. The reverse dynamics are parameterized by the EDM noise variable $\sigma$, rather than by the variance-preserving time variable $t$ used in the analytic patch model.

The deterministic reverse flow is written as
\[
\frac{dx}{d\sigma} = \driftsigma(x,\sigma),
\qquad
\driftsigma(x,\sigma)=\frac{x-D_\theta(x,\sigma,y)}{\sigma},
\]
where $D_\theta(x,\sigma,y)$ denotes the class-conditional denoiser and $y$ is the class label.

The guidance-type denoiser is
\[
D_{\rm cfg}(x,\sigma,y;w)
=
D_\theta(x,\sigma,y)
+
w\Bigl(D_\theta(x,\sigma,y)-D_{\rm u}(x,\sigma)\Bigr),
\]
where $w$ is the guidance scale.

The corresponding reverse drift under guidance is therefore
\[
b_\sigma^{(w)}(x,\sigma)
=
\frac{x-D_{\rm cfg}(x,\sigma,y;w)}{\sigma}.
\]

Sampling is performed using the standard EDM sampler with Heun correction, with noise levels evolving from
$\sigma_{\max}=80$
to
$\sigma_{\min}=0.002$.

The noise schedule is
\[
\sigma_i =
\left(
\sigma_{\max}^{1/\rho}
+
\frac{i}{N-1}
\left(\sigma_{\min}^{1/\rho}-\sigma_{\max}^{1/\rho}\right)
\right)^\rho
\]
with
$\rho = 7$.

\subsection{Dynamical diagnostic used for pulse placement}
The reported intervention uses centered luminance, $\widetilde\ell=\ell-\langle\ell\rangle$, and the continuous-field correlation
\[
 C(r;\sigma)=\langle\widetilde\ell(z)\widetilde\ell(z+r)\rangle.
\]
Its radial second-moment statistic has the form
\[
 \xi_x(\sigma)=\left[\frac{\sum_{r\in\mathcal R}r^2C(r;\sigma)}{\sum_{r\in\mathcal R}C(r;\sigma)}\right]^{1/2},
\]

We estimate Fourier-direction Jacobian responses by symmetric finite differences of the unguided drift and fit a normalized low-frequency progression,
\[
 \lambda_v(\sigma)\simeq
 \frac{v^\top[\driftsigma(x+\epsilon v,\sigma)-\driftsigma(x-\epsilon v,\sigma)]}{2\epsilon\,v^\top v},
 \qquad
 \frac{\lambda_n}{\lambda_0}\simeq1+\xi_{\rm eq}^2(nk_{\min})^2.
\]
As in the unguided analysis, this defines a dispersion proxy rather than an equilibrium observable. 

\subsection{Global critical scale estimation}

A naive strategy would be to estimate a separate critical scale for each class. In practice, however, this estimate is noisy because the Jacobian-based equilibrium-length proxy is sensitive to finite-sample fluctuations.

To reduce variance, we used a global procedure. We first generated unguided trajectories for all selected classes and computed $\xi_{\rm eq}(\sigma)$ for each trajectory. These curves were then aggregated across all classes and trajectories, and a single global critical scale was defined as
\[
\sigma_c = \arg\max_\sigma \mathrm{median}\,\xi_{\rm eq}(\sigma).
\]

This construction has two advantages:

\begin{itemize}
\item it suppresses class-specific noise in the equilibrium-length estimate,
\item it identifies a single intervention scale that can be applied uniformly across classes.
\end{itemize}

Operationally, $\sigma_c$ is the maximum of an aggregated, processed dispersion proxy; this definition does not prove that every low-frequency mode or class is maximally soft at the same scale.

\subsection{Pulse-guidance intervention}

After estimating the global critical scale, we ran two intervention conditions.

\paragraph{Critical-pulse condition.}
A guidance pulse of strength $w_{\rm pulse}$ was applied only within a short interval centered at the global critical scale:
\[
w(\sigma)=
\begin{cases}
w_{\rm pulse}, & \sigma \in [\sigma_c-\Delta,\sigma_c+\Delta],\\
0, & \text{otherwise}.
\end{cases}
\]

\paragraph{Random-pulse condition.}
A pulse of identical strength and width was applied, but with center sampled uniformly at random in $\sigma$:
\[
\sigma_{\rm rand}\sim \mathrm{Unif}[\sigma_{\min},\sigma_{\max}],
\]
and
\[
w(\sigma)=
\begin{cases}
w_{\rm pulse}, & \sigma \in [\sigma_{\rm rand}-\Delta,\sigma_{\rm rand}+\Delta],\\
0, & \text{otherwise}.
\end{cases}
\]

The prescribed conditions differ in pulse center. Equal continuous width does not ensure equal realized exposure on a discrete schedule: active steps and endpoint clipping must be checked before treating the interventions as dose matched.

\subsection{Classes, trials, and hyperparameters}

We selected 10 ImageNet classes and generated 40 samples per class and per condition.

The pulse experiment therefore consisted of
\[
10 \text{ classes} \times 2 \text{ conditions} \times 40 \text{ trials}
\]
for a total of 800 guided generations.

The key hyperparameters were:

\begin{center}
\begin{tabular}{ll}
Number of classes & 10 \\
Trials per class per condition & 40 \\
Critical scale & global, estimated from all classes \\
Pulse width & $2\Delta$ in $\sigma$ \\
Pulse strength & fixed strong guidance scale \\
Conditions & critical pulse / random pulse
\end{tabular}
\end{center}

\subsection{Class-alignment metric}

To quantify whether the pulse improved semantic targeting, final generated images were evaluated with a pretrained DINOv2 ImageNet classifier.

For a generated image $x$, the classifier produces logits $z_c(x)$ and probabilities
\[
p_c(x)=\frac{e^{z_c(x)}}{\sum_{c'} e^{z_{c'}(x)}}.
\]

For a target class $y$, we recorded:
\begin{itemize}
\item the target-class logit $z_y(x)$,
\item the target-class probability $p_y(x)$,
\item whether the top predicted class matched the target class.
\end{itemize}

These quantities were averaged across trials for each class and condition.

\subsection{Statistical comparison}

For each class, we compared the critical-pulse and random-pulse conditions using the mean target probability, mean target logit, and top-1 correctness fraction. The per-class effect summaries are defined by

\[
\Delta p_y = \bar p_y^{\rm critical} - \bar p_y^{\rm random},
\]
\[
\Delta z_y = \bar z_y^{\rm critical} - \bar z_y^{\rm random}.
\]

Positive values indicate that the pulse delivered near the critical scale yields better class alignment than a pulse delivered at a random point.

\subsection{Interpretation}

The theoretical picture developed in the main text predicts that the reverse diffusion dynamics pass through a critical window in which long-wavelength modes are especially soft. In that regime, perturbations to the drift should have disproportionate influence on the macroscopic organization of the sample.

The comparison tests whether the prescribed critical-scale pulse improves class alignment relative to the prescribed random-center pulse. It does not alone establish a maximum of susceptibility or isolate critical softening from other timing-dependent effects.

The reported class-alignment differences support the operational relevance of intervention timing under this protocol. Stronger claims of optimality or mechanism require a response scan and controls for realized pulse exposure.

\subsection{Visualization of guided and unguided trajectories}

To complement the quantitative metrics, we also visualized denoising trajectories under unguided sampling and under critical-pulse sampling. For a fixed class and random seed, we plotted a sequence of intermediate denoising states across 10 logarithmically spaced values of $\sigma$.

These visualizations compare trajectories initialized with the same seed. Outside the pulse, the prescribed guidance rule is unchanged, but the altered state can affect all subsequent denoising steps. The images illustrate that propagation; they do not isolate a local causal mechanism by themselves.

% BEGIN KZ SUPPLEMENT ADDITION
\section{Supplementary theory and experiments: quantitative Kibble--Zurek domain seeding}
\label{supp:kz}
\subsection{Protocol, notation, and scope}
We specialize the analytic score to the two equally probable ferromagnetic patches, with $\Omega=\{-K,\ldots,K\}^2$ and $Q=|\Omega|=(2K+1)^2$. This is the model underlying Eqs.~\eqref{eq:patch_expansion}--\eqref{eq:patch_lambda_exact}, rather than the mixed dictionary used in the preceding patch experiments. All lattice spacings are one. The symbol $x$ remains the continuous noisy field, $\binfield=\operatorname{sign}(x)$ its binarization, and $\lambda$ an eigenvalue of $\RevJac$, not of the oppositely signed $\Jac$.

For $\beta=1$, the score coefficients are evaluated at $\alpha(t)=e^{-t/2}$ and $\sigma^2(t)=1-e^{-t}$. We introduce the controlled dynamics
\begin{equation}
 dx_i=\left[\frac12 x_i+\frac{\alpha(t)}{\sigma^2(t)}
 \tanh\!\left(\frac{\alpha(t)}{\sigma^2(t)}\sum_{u\in\Omega}x_{i+u}\right)
 -\frac{x_i}{\sigma^2(t)}\right]d\tau+\sqrt{2D}\,dW_i,
 \qquad t=t_c-\gamma(\tau-\tau_c).
 \label{eq:kz-controlled-sde}
\end{equation}
Here the clock $\tau$ has the same direction and drift normalization as the denoising clock in the main text, but the independent schedule is slowed. For $\gamma\ne1$ this is a modified dynamics, not a reparameterization of the native reverse SDE: a mere clock change would rescale both drift and noise. The native $\beta=1$ sampler corresponds to $\gamma=1$, $D=1/2$. Weak noise is used to test the Gaussian prediction before substantial fluctuation renormalization. The exact mean-field $t_c$ from Appendix~\ref{app:gl-critical-time}, rather than $t_c^{\rm approx}$ in the original plots, is essential for a quantitative comparison.

\subsection{The crossing seed scale and the full critical-window spectrum}
Let $v_0=r'(t_c)$, $\kappa_c=\kappa(t_c)$ and $u_c=u(t_c)$. For $a_c=e^{-t_c}$ and $s_c=1-a_c$, the derivative is
\begin{equation}
 v_0=-\frac{a_c}{s_c^2}+\frac{Q a_c}{s_c^2}+\frac{2Q a_c^2}{s_c^3}.
\end{equation}
For $K=2$ this gives $t_c=3.9124228$, $v_0\simeq0.520816$, $\kappa_c\simeq0.520400$, and $u_c\simeq2.25680$. Near the crossing, $r(t)=-v(\tau-\tau_c)+O(\gamma^2(\tau-\tau_c)^2)$, where $v=\gamma v_0$. Thus, before nonlinear saturation,
\begin{equation}
 \partial_\tau x_k=\big[v(\tau-\tau_c)-\kappa_c|k|^2\big]x_k+\eta_k,
 \qquad \langle\eta_i(\tau)\eta_j(\tau')\rangle=2D\delta_{ij}\delta(\tau-\tau').
\end{equation}
The zero-mode relaxation time $1/|r|$ becomes comparable to the time to the crossing at $\hat\tau=v^{-1/2}$; the corresponding equilibrium length $\sqrt{\kappa_c/|r|}$ gives $\hat\xi=\sqrt{\kappa_c}v^{-1/4}$. This freeze-out construction gives the rate exponent but does not alone determine a domain diameter.

Use an orthonormal discrete Fourier transform, so $S(k,\tau)=\mathbb E|x_k|^2$ has equilibrium value $D/[-\lambda(k,t)]$ on the stable branch. For an initialization sufficiently far on that branch, integrating the linear stochastic equation gives
\begin{align}
 S(k,\tau)&=2D\int_{-\infty}^{\tau-\tau_c}
 \exp\!\left\{v[(\tau-\tau_c)^2-z^2]-2\kappa_c|k|^2[(\tau-\tau_c)-z]\right\}dz,\\
 F_0(q,y):=\frac{\sqrt v}{D}S(k,y)
 &=\sqrt\pi\,\operatorname{erfcx}(q^2-y),\qquad q=|k|\hat\xi.
 \label{eq:kz-supp-spectrum}
\end{align}
In particular $S(0,0)=D\sqrt{\pi/v}$, while the curvature at $k=0$ gives
\begin{equation}
 \xi_{\rm spec}^2(y):=-\left.\partial_{|k|^2}\log S(k,y)\right|_{k=0}
 =\hat\xi^2\left[2y+\frac{2}{\sqrt\pi\operatorname{erfcx}(-y)}\right].
 \label{eq:kz-supp-length}
\end{equation}
Consequently $\xi_{\rm spec}(0)=\xi_\times^{\rm KZ}=1.06225\hat\xi$. This observable is different from $\xi_x$, the first moment of the positive radial autocorrelation of $\binfield$ in the original experiments. Even in a Gaussian field, thresholding changes the correlation function: for a centered unit-variance pair with correlation $c$, the sign correlation is $(2/\pi)\arcsin c$. There is therefore no justification for equating those two lengths numerically. At nonzero $y$, Eq.~\eqref{eq:kz-supp-length} predicts a changing spectral length; freeze-out is not a literal plateau in every measured length.

\subsection{Controlled asymptotic window}
The conditions $K\ll\hat\xi\ll L$ and weak nonlinear fluctuations require, parametrically in two dimensions,
\begin{equation}
 g_{\rm KZ}=\frac{D u_c}{\kappa_c\sqrt{\gamma v_0}}\ll1,\qquad
 \max\!\left\{\frac{\kappa_c^2}{v_0L^4},\frac{1}{v_0}\left(\frac{Du_c}{\kappa_c}\right)^2\right\}
 \ll\gamma\ll\frac{\kappa_c^2}{v_0K^4}.
\end{equation}
The fluctuation criterion is a scale estimate, not a rigorous bound; microscopic contributions can introduce additional logarithmic dependence in two dimensions. With $K=2$, $D=2\times10^{-5}$ and the four rates below, $g_{\rm KZ}\simeq0.012$--$0.038$, and $\hat\xi/K\simeq4.25$--$7.55$. The regime is deliberately different from native sampling. Predictions at $D=1/2$ cannot be inferred by increasing the noise amplitude in the Gaussian formulas without checking renormalization and nonlinear saturation.

\subsection{Exact lattice covariance tests}
We propagate the deterministic covariance equation
\begin{equation}
 \frac{dS(k,\tau)}{d\tau}=2\lambda(k,t)S(k,\tau)+2D
\end{equation}
with the exact Fourier symbol in Eq.~\eqref{eq:patch_lambda_exact}. Initialization at $y=-6$ uses the stationary covariance $-D/\lambda(k,t)$; the zero-mode memory at the crossing is suppressed at Gaussian order by $e^{-36}$. An adaptive integrator uses relative tolerance $2\times10^{-9}$ and absolute tolerance $10^{-12}$. To avoid conflating finite image size with finite rate, the length in Table~\ref{tab:kz-fresh-rates} is the intrinsic curvature: we also propagate $P=\partial_{k_x^2}S|_0$ through
\begin{equation}
 \partial_\tau P=2\lambda(0,t)P-2\kappa(t)S(0,\tau),\qquad
 \xi_{\rm spec}^2=-P/S(0).
\end{equation}
This does not require a fit to discrete low-frequency modes. The spectra in the deterministic covariance tests use wavevectors along a lattice axis; lattice anisotropy away from small $k$ is retained in the numerical symbol.

\begin{table}[t]
\centering
\caption{Exact-lattice covariance tests at $K=2$, $D=2\times10^{-5}$. The exact length is the intrinsic spectral curvature, without a finite-$L$ estimator.}
\label{tab:kz-fresh-rates}
\begin{tabular}{rrrrr}
\toprule
$\gamma$ & $\hat\xi$ & $\xi_\times^{\rm KZ}$ & $\xi_\times^{\rm exact}$ & $S(0,0)$ \\
\midrule
1e-04 & 8.492 & 9.020 & 9.009 & 0.004926 \\
5e-05 & 10.098 & 10.727 & 10.717 & 0.006961 \\
2e-05 & 12.698 & 13.489 & 13.481 & 0.010998 \\
1e-05 & 15.101 & 16.041 & 16.034 & 0.015547 \\
\bottomrule
\end{tabular}
\end{table}

\subsection{Full nonlinear ensembles: independent tests of seed sizes}
The nonlinear ensemble experiment uses $K=2$, $L=128$, $D=2\times10^{-5}$ and $64$ independent full-score trajectories at each of $\gamma=10^{-4},4\times10^{-5},10^{-5}$. A separate $32$-trajectory challenge uses $D=2\times10^{-3}$ at $\gamma=10^{-4}$. Euler--Maruyama uses $\Delta\tau=0.4$. Initialization at $y=-4$ samples the stationary covariance of the exact lattice linearization, and each nonlinear realization has a linear partner initialized identically and driven by the same Wiener increments. Measurements occur at $y=-1,0,1,1.5,2,3,4$, shortening the final step to hit each time exactly. The independent seeds are $202610121$, $202610122$, $202610123$ and $202610124$ for the fastest, intermediate, slowest and stronger-noise conditions, respectively.

There are two different uses of the spectra. To \emph{measure} a seed length, we fit $A\operatorname{erfcx}(|k|^2\ell^2)$ at the crossing, allowing both $A$ and $\ell$ to vary. The reported length is $\xi_{\rm fit}=\sqrt{2/\sqrt\pi}\ell$. The zero mode is excluded, and a fixed physical cutoff $0<|k|\le0.25$ is used at every rate. Thus the theoretical rate dependence is not imposed by the estimator. Powers are averaged on rings of equal $|k|^2$, and the residual is the log spectral difference weighted by the square root of the ring mode count. This is a model-based shape estimate, not a model-independent curvature measurement; its interpretation depends on the adequacy of the Gaussian shape. The stronger \emph{prediction test} compares the full nonlinear spectrum with the independently specified shape and amplitude in Eq.~\eqref{eq:kz-supp-spectrum}, without fitting either quantity.

Uncertainty is estimated by $400$ bootstrap resamples of whole trajectories, with fixed seeds $813$--$816$ for the length estimates. Each resample preserves dependencies among Fourier modes and their conjugates. We report $95\%$ percentile intervals. Independent condition-wise bootstrap length draws also give an interval for the log--log rate slope. The fitted exponent is $-0.2453$, with $95\%$ interval $[-0.3441,-0.1701]$. This supports consistency with $-1/4$ at the available precision; it cannot distinguish nearby exponents or establish a universality class.

\begin{table}[t]
\centering
\caption{Crossing seed-scale tests in the full nonlinear score, $64$ trajectories per weak-noise rate. The measured length fits amplitude and shape independently; bracketed values are $95\%$ trajectory-bootstrap intervals. The matched linear estimate uses the same realizations and noise. $E_0$ is the no-fit mean absolute shell deviation, and $\delta_0$ is the mean paired RMS difference divided by the mean nonlinear field RMS, at $y=0$.}
\label{tab:kz-nonlinear-seeds}
\small
\begin{tabular}{@{}rrrrrr@{}}
\toprule
$\gamma$ & $\xi_\times^{\rm KZ}$ & $\xi_{\rm fit}$ [interval] & $\xi_{\rm fit}^{\rm linear}$ & $E_0$ (\%) & $\delta_0$ (\%) \\
\midrule
1e-04 & 9.02 & 9.02 [8.25, 9.91] & 9.07 & 5.12 & 0.170 \\
4e-05 & 11.34 & 10.09 [9.25, 11.26] & 10.21 & 4.68 & 0.290 \\
1e-05 & 16.04 & 15.64 [13.64, 19.01] & 16.12 & 9.49 & 0.673 \\
\bottomrule
\end{tabular}
\end{table}

At the middle rate, the prediction lies just above the nominal percentile interval for the measured length, while its paired linear estimate has a similar deficit. The same-noise nonlinear/linear fitted-length differences are about $0.5\%$, $1.1\%$ and $3.0\%$ at increasing slowness. These observations caution against treating every finite-ensemble length discrepancy as a nonlinear breakdown. The bootstrap intervals quantify sampling variability of the estimator, not its possible model bias. The complete no-fit spectral comparison is therefore the primary predictive check.

\subsection{No-fit spectra, paired controls and the breakdown regime}
For each condition, annuli of width $\Delta q=0.3$ cover $0.4\le q<2.5$; empty annuli are omitted. Every theoretical shell is averaged over exactly the same discrete Fourier modes as the data. Let $\overline S_b$ denote the ensemble and shell average and $\overline S_b^{\rm KZ}$ its prediction. The plotted spectral diagnostic is
\begin{equation}
 E(y)=\frac1{N_b}\sum_b\left|\frac{\overline S_b(y)}{\overline S_b^{\rm KZ}(y)}-1\right|.
\end{equation}
The $95\%$ shading in Fig.~\ref{fig:kz-domain-prediction}c is a trajectory-bootstrap interval for this diagnostic, using seeds $918$--$921$. It has a positive sampling-noise bias, so it is not a calibrated goodness-of-fit statistic or a $p$-value. Spectral point bars in panels (b,e) instead show $1.96$ ensemble standard errors. In weak noise, $E$ remains approximately $5$--$10\%$ through $y=1.5$ and reaches $35$--$50\%$ by $y=3$ (Table~\ref{tab:kz-expanded-errors}).

\begin{table}[t]
\centering
\caption{Mean absolute no-fit shell deviations $100E(y)$ in percent. The first three columns use $D=2\times10^{-5}$ and $64$ trajectories each; the final column uses $D=2\times10^{-3}$ and $32$ trajectories at $\gamma=10^{-4}$. These positive deviations include sampling noise.}
\label{tab:kz-expanded-errors}
\begin{tabular}{rrrrr}
\toprule
$y$ & $\gamma=10^{-4}$ & $\gamma=4\times10^{-5}$ & $\gamma=10^{-5}$ & Stronger noise \\
\midrule
-1 & 4.7 & 5.1 & 9.0 & 14.8 \\
0 & 5.1 & 4.7 & 9.5 & 25.5 \\
1 & 4.9 & 6.4 & 7.8 & 44.8 \\
1.5 & 4.8 & 7.6 & 6.7 & 53.0 \\
2 & 7.6 & 10.8 & 14.3 & 62.3 \\
3 & 35.3 & 39.2 & 49.5 & 80.3 \\
4 & 207.4 & 134.9 & 71.9 & 91.9 \\
\bottomrule
\end{tabular}
\end{table}

The paired controls have a different purpose: identical noise cancels much of the sampling scatter and isolates the nonlinear correction to the trajectory. Define
\begin{equation}
 \delta(y)=\frac{\mathbb E[\|x^{\rm nonlinear}-x^{\rm linear}\|_{\rm RMS}]}{\mathbb E[\|x^{\rm nonlinear}\|_{\rm RMS}]}.
\end{equation}
At weak-noise crossing, $\delta=0.00170,0.00290,0.00673$ for the three rates. By $y=3$, it exceeds one in all three conditions: the unsaturated linear field has grown beyond the nonlinear field. This ratio can be arbitrarily large; it is not bounded between zero and one. Panel (f) shows that the late discrepancy is wavelength dependent: saturation suppresses the longest-wavelength linear growth, while domain interfaces introduce additional shorter-wavelength power. A single multiplicative amplitude correction is consequently insufficient.

At stronger noise, $g_{\rm KZ}\simeq1.2$ rather than $0.012$ at the fastest weak-noise rate. Already at the crossing, $E\simeq25.5\%$ and $\delta\simeq0.101$. Fitting the Gaussian shape gives $\xi_{\rm fit}\simeq6.10$ with interval $[5.26,6.97]$, compared with the predicted $9.02$ and the paired linear estimate $8.36$. This shape-derived length is exploratory because the Gaussian approximation is now less accurate. The suppressed long-wavelength spectrum and paired discrepancy directly demonstrate a loss of predictive validity. The comparison is consistent with fluctuation-induced mass corrections, but does not determine a renormalized critical point, an interacting critical exponent, or the native-sampler behavior.

\subsection{Numerical refinement and reproducibility}
An independent eight-trajectory Brownian-coupled refinement uses seed $202610129$ at $\gamma=10^{-4}$ and $D=2\times10^{-5}$. Two fine-step Wiener increments at $\Delta\tau=0.2$ sum to the increment of the corresponding $0.4$ step. Thus the coarse and fine nonlinear integrators see the same noise path. Mean absolute shell changes are $0.141\%$ at the crossing and $0.157\%$ at $y=1.5$; the largest shell change is $0.328\%$. The crossing shape length changes by $0.182\%$. A corresponding stronger-noise refinement with seed $202610130$ gives mean shell changes $0.215\%$ and $0.338\%$ at $y=0,1.5$, respectively, and a crossing fitted-length change of $0.453\%$. Both checks support using $\Delta\tau=0.4$ for the early-window comparisons, without asserting convergence of every late-time observable.

A direct finite-image estimator sometimes used is
\begin{equation}
 \xi_{\rm fd}^2=\frac{S(0)/\overline S(k_{\min})-1}{k_{\min}^2},\qquad k_{\min}=2\pi/L.
\end{equation}
It has a relative finite-grid bias $O[(\hat\xi/L)^2]$ and can become negative under sampling fluctuations because of the noisy zero mode. The intrinsic linear lengths and the finite-grid-aware spectral shape estimates here avoid treating that estimator as the sole test of the prediction.

\subsection{Receptive-field size and predicted domain seeds}
\begin{figure}[t]
\centering
\includegraphics[width=.85\linewidth]{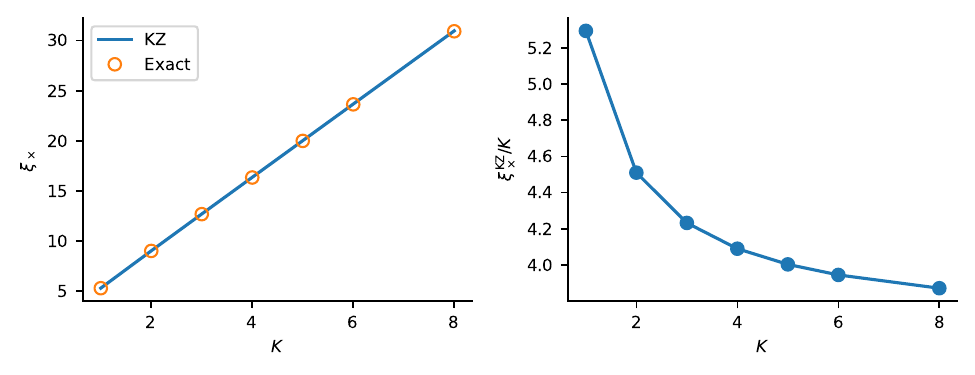}
\caption{Receptive-field dependence at fixed $\gamma=10^{-4}$. Left: predicted crossing scale and exact intrinsic lattice curvature. Right: the relative separation from the patch radius approaches a constant, despite the increasing absolute length.}
\label{fig:kz-window-size}
\end{figure}

Changing $K$ changes the coefficients as well as the microscopic interaction radius. In two dimensions, as $K\to\infty$,
\begin{equation}
 t_c\sim\log(8K^2),\qquad v_0\to\tfrac12,\qquad
 \kappa_c\sim K^2/12,\qquad u_c\sim K^2/3.
\end{equation}
At fixed $\gamma$, $\xi_\times^{\rm KZ}$ therefore grows proportionally to $K$, but
\begin{equation}
 \frac{\xi_\times^{\rm KZ}}{K}\longrightarrow
 \frac{\sqrt{2/\sqrt\pi}}{\sqrt{12}}\left(\frac2\gamma\right)^{1/4},\qquad
 g_{\rm KZ}\longrightarrow\frac{4\sqrt2D}{\sqrt\gamma}.
\end{equation}
Thus a larger window increases absolute structure sizes but does not by itself improve the separation from the receptive-field radius or suppress fluctuation effects. Our independent intrinsic-curvature calculation at $\gamma=10^{-4}$ for $K=1,2,3,4,5,6,8$ tests this trend (Fig.~\ref{fig:kz-window-size}); exact values are included in \texttt{kz\_windows.csv}. For image experiments, keeping $L/K$ fixed is necessary to separate this effect from finite-size saturation.

\subsection{Corrections to scaling and finite-size effects}
The critical interval in control space has width $\epsilon=\gamma\hat\tau=\sqrt{\gamma/v_0}$. Taylor expansion of the exact coefficients and lattice symbol gives
\begin{equation}
 \frac{\lambda(k,t)}{\sqrt v}=y-q^2+\epsilon H_K(q,y)+O(\epsilon^2),\qquad
 H_K=A_Ky^2+B_Kyq^2+C_K(\widehat k)q^4,
\end{equation}
where $A_K=-r''(t_c)/(2v_0)$, $B_K=\kappa'(t_c)/\kappa_c$, and
$C_K(\widehat k)=v_0c_4(\widehat k,t_c)/\kappa_c^2$. The direction dependence is essential for a square patch. For a unit vector $n=\widehat k$, define $m_2=K(K+1)/3$ and $m_4=K(K+1)(3K^2+3K-1)/15$. The fourth-order coefficient is
\begin{equation}
 c_4(n,t_c)=\frac{a_cQ}{s_c^2}
 \left[\frac{m_4}{24}(n_x^4+n_y^4)+\frac{m_2^2}{4}n_x^2n_y^2\right].
\end{equation}
The rescaled covariance $F=\sqrt vS/D$ consequently has expansion $F=F_0+\epsilon F_1+O(\epsilon^2)$ with
\begin{align}
 \partial_y F_1&=2(y-q^2)F_1+2H_K F_0,\\
 F_1(q,y)&=2\int_{-\infty}^y
 e^{y^2-z^2-2q^2(y-z)}H_K(q,z)F_0(q,z)\,dz.
\end{align}
This integral is an asymptotic critical-window construction: matching to an early stable regime removes initial-condition transients. At the crossing the first-order correction to the length follows from
\begin{equation}
 \frac{\xi_\times^{\rm exact}}{\xi_\times^{\rm KZ}}
 =1+\epsilon c_K+O(\epsilon^2),\qquad
 c_K=\frac12\left[\frac{\partial_{q^2}F_1(0,0)}{\partial_{q^2}F_0(0,0)}-\frac{F_1(0,0)}{F_0(0,0)}\right].
\end{equation}
Intrinsic finite-rate corrections vanish as $\sqrt\gamma$, whereas the finite-image curvature bias grows as $\gamma^{-1/2}/L^2$. They must be separated before interpreting an effective exponent. The reproducible intrinsic-curvature experiment here avoids the latter; a finite-grid effective exponent should not be identified with a changed universality exponent without this separation. The functional expansion provides a theoretical residual-collapse test, but the crossing-scale and spectral tests reported here do not by themselves establish numerical convergence to $F_1$.

\subsection{From critical correlations to visible domains}
The bars in Fig.~\ref{fig:kz-domain-prediction}g--i mark a prediction fixed before observing the trajectory. At the crossing it refers to a continuous-field spectral correlation scale, not the diameter of every connected sign region. Weak microscopic fluctuations remain present, so the raw sign field is not a direct geometrical ruler. After the crossing, low-frequency instability amplifies the seeded fluctuations. A useful estimate of the onset of nonlinear effects is $u_c\mathbb E[x^2]\sim\sqrt v\,y$; this is not an exact saturation-time criterion for the full tanh dynamics. At later times domain growth may become approximately curvature driven, $L_{\rm dom}^2\simeq L_{\rm onset}^2+C\kappa_c(\tau-\tau_{\rm onset})$. Equivalently, $L_{\rm dom}^2/\hat\xi^2\simeq (L_{\rm onset}/\hat\xi)^2+C(y-y_{\rm onset})$. Neither the nonuniversal coefficient $C$ nor $L_{\rm onset}$ is fixed solely by $\xi_\times^{\rm KZ}$. Agreement of a later sign-domain length with a guessed $C=1$ would therefore not constitute a parameter-free KZ test. The strongest quantitative evidence is the independently predicted crossing scale together with the critical-window spectrum, while the snapshots illustrate their role in subsequent spatial organization.

\subsection{Nonlinear wall motion beyond the critical window}
\label{supp:domain-walls}
The spectral prediction determines the correlations available to seed domains, but does not determine their subsequent geometry. We therefore test a separate reduction of the nonlinear field equation. These tests retain the two equally weighted ferromagnetic patches with $K=2$ and use the full lattice tanh score in Eq.~\eqref{eq:kz-controlled-sde} as the reference dynamics. They concern the analytic patch model, not a trained neural score.

\subsubsection{A leading wall model.}
For $r(t)<0$, write $a(\tau)=-r(t(\tau))>0$. Freezing the coefficients momentarily, the cubic field equation has bulk values $x_\pm=\pm\sqrt{a/u}$ and a planar wall
\begin{equation}
 X(n)=\sqrt{a/u}\tanh(n/\ell),\qquad
 \ell=\sqrt{2\kappa/a}.
 \label{eq:wall-profile}
\end{equation}
If $n$ increases outward from a domain and $\mathcal K=\nabla\cdot\mathbf n$ is positive on a convex boundary, substituting $\nabla^2X\simeq X''+\mathcal KX'$ and $\partial_\tau X\simeq-V_nX'$ gives
\begin{equation}
 V_n=-\kappa(t)\mathcal K,\qquad
 R^2(\tau)-R^2(\tau_0)=-2\int_{\tau_0}^{\tau}\kappa(t(s))\,ds
 \label{eq:wall-curvature}
\end{equation}
for a circular domain in two dimensions. A small sinusoidal displacement of a planar wall, of wave number $k$, decays at rate $\kappa k^2$ for fixed coefficients. Neither prediction fits a wall-motion coefficient. They require weak fluctuations, walls thin compared with their radius of curvature, and a profile that adjusts faster than the drive. The cubic profile additionally requires small field amplitudes, while a continuum wall requires its thickness to exceed the patch scale. Scale separation is only moderate in the finite tests below, so agreement is numerical evidence rather than an asymptotic convergence demonstration. These conditions need not hold throughout native denoising.

\subsubsection{Keeping nonlinear amplitude dependence.}
A second reduction expands only the patch convolution and retains the tanh response. Set $c(t)=\alpha(t)/\sigma^2(t)$ and $Q=25$. To second order in spatial derivatives,
\begin{align}
 \partial_\tau x&=f(x,t)+B(x,t)\nabla^2x,\\
 f(x,t)&=\big[\tfrac12-1/\sigma^2(t)\big]x+c(t)\tanh(c(t)Qx),\\
 B(x,t)&=\kappa(t)\operatorname{sech}^2(c(t)Qx).
 \label{eq:wall-full-tanh}
\end{align}
This preserves the exact homogeneous drift, but remains a spatial-gradient approximation. A planar stationary wall satisfies $BX''+f=0$. The linearized wall operator is $\mathcal L=B\partial_n^2+[\partial_x f+(\partial_x B)X'']$. Differentiating the profile equation gives $\mathcal L X'=0$, while its adjoint has null mode $X'/B$. Projecting $-V_nX'=B\mathcal KX'$ onto that adjoint mode gives
\begin{equation}
 V_n=-\kappa_{\rm wall}(t)\mathcal K,\qquad
 \kappa_{\rm wall}=
 \frac{\int (X')^2\,dn}{\int (X')^2/B(X,t)\,dn}.
 \label{eq:wall-mobility}
\end{equation}
We calculate this coefficient without observing moving walls. If $x_b>0$ is the exact homogeneous root of $f(x_b,t)=0$, the planar equation gives
\begin{equation}
 p(x)^2=2\int_x^{x_b}\frac{f(z,t)}{B(z,t)}\,dz,\qquad
 \kappa_{\rm wall}=
 \frac{\int_0^{x_b}p(x)\,dx}{\int_0^{x_b}p(x)/B(x,t)\,dx},
 \label{eq:wall-quadrature}
\end{equation}
where $p=X'$ on the increasing wall. The resulting mobility approaches $\kappa$ near the transition and decreases as nonlinear polarization becomes important. It does not justify extrapolating the gradient expansion to walls at the lattice scale.

\subsubsection{Controlled geometries.}
\begin{figure}[t]
 \centering
 \includegraphics[width=\textwidth]{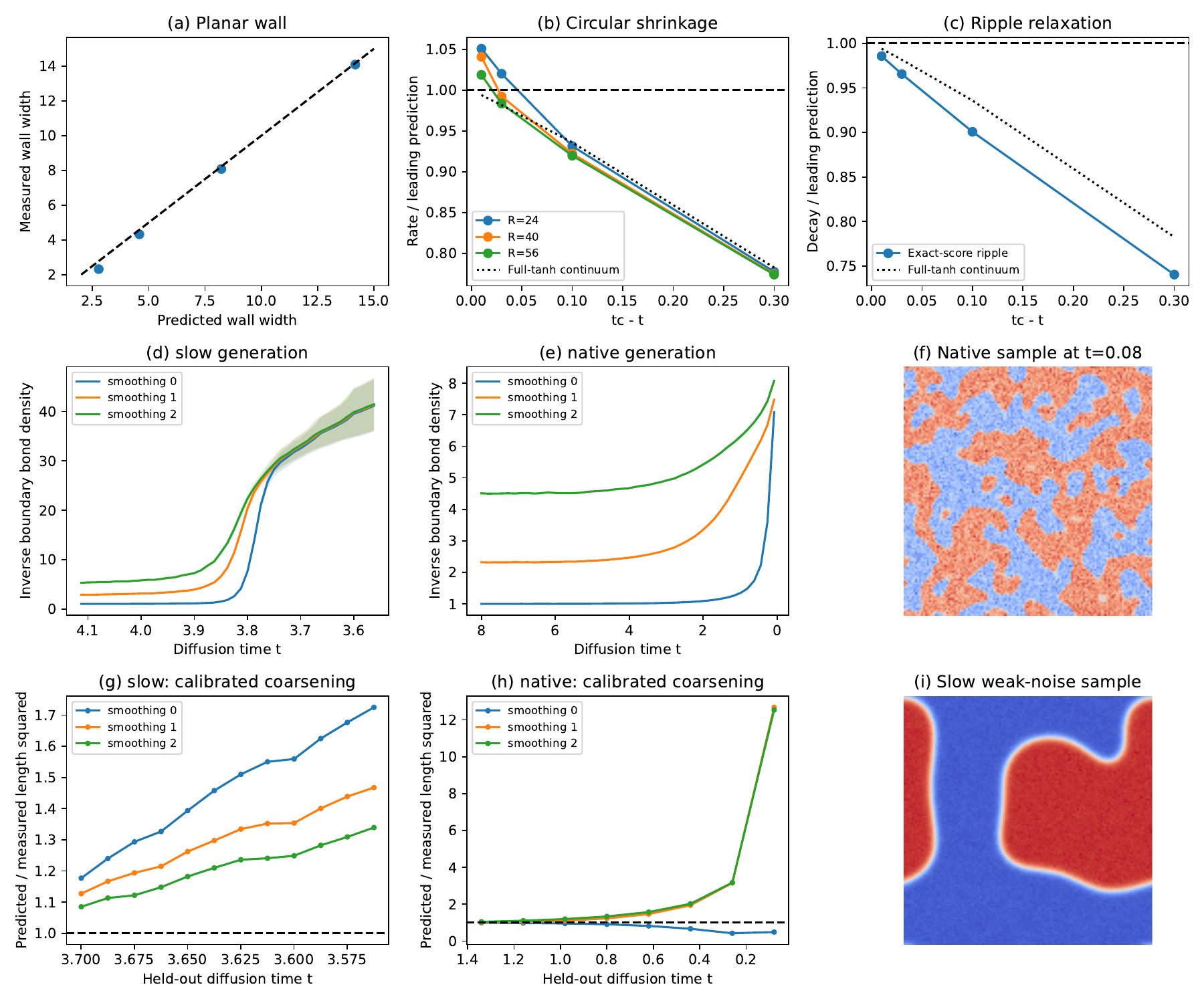}
 \caption{\textbf{Controlled wall tests and stochastic domain formation.} (a) Planar width estimates against the cubic prediction. (b) Circular shrinkage relative to the leading curvature law, at three radii. (c) Sinusoidal-wall decay relative to the leading prediction. Dotted curves in (b,c) show the independent full-tanh continuum mobility. (d,e) Inverse sign-boundary bond density during slow weak-noise and native generation, with Gaussian smoothing of width $0,1,2$ sites; bands are $1.96$ trajectory standard errors. (f,i) Representative terminal fields. (g,h) Later measured squared lengths compared with a coarsening law calibrated on earlier observations. These morphology checks have fitted coefficients and must be distinguished from the controlled no-fit predictions.}
 \label{fig:wall-controlled}
\end{figure}

With $D=0$, at $t_c-t=0.01,0.03,0.1,0.3$, we relax periodic planar slabs on a $192^2$ lattice and test circles of initial radius $24,40,56$. The leading wall-width parameters are $14.17,8.22,4.58,2.78$ sites; the complete measurements are recorded with the experiments. Near the transition, circular shrinkage rates agree with $-2\kappa$ to within $5.1\%$ and the decay of a wavelength-$96$, amplitude-$3$ wall displacement agrees with $\kappa k^2$ to within $3.5\%$. At $t_c-t=0.3$, circular and ripple motion are approximately $22\%$ and $26\%$ slower than the leading predictions. Equation~\eqref{eq:wall-quadrature} predicts a $21.7\%$ reduction in mobility, explaining much, but not all, of the discrepancy (Fig.~\ref{fig:wall-controlled}).

\begin{figure}[t]
 \centering
 \includegraphics[width=\textwidth]{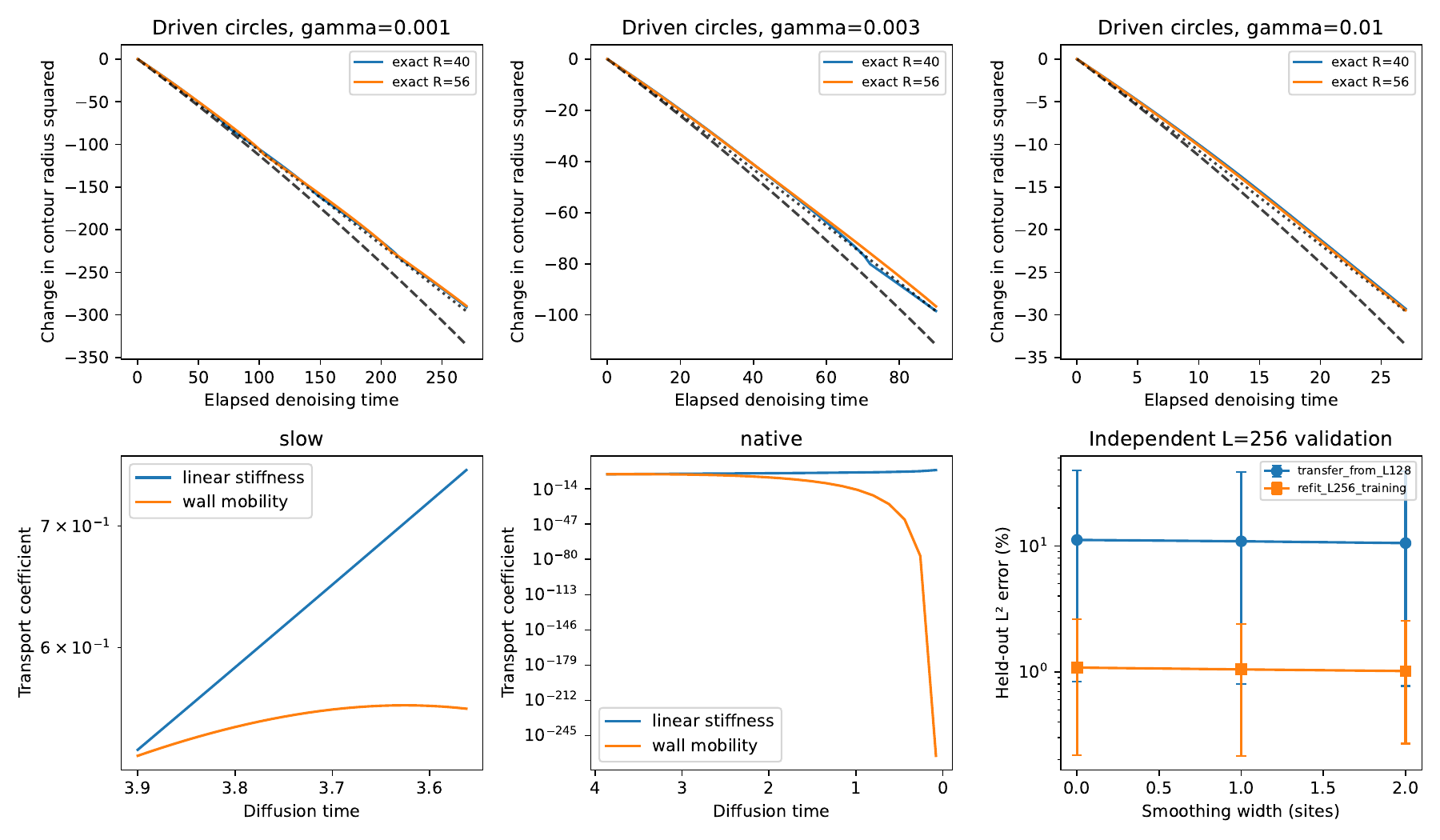}
 \caption{\textbf{Driven circles and limits of network extrapolation.} Top: changes in zero-contour squared radius for two initial radii and three ramp rates; solid curves are full-score dynamics, dashed curves integrate the leading stiffness, and dotted curves integrate the independently calculated full-tanh wall mobility. Bottom left and center: linear stiffness and nonlinear wall mobility along the slow and native schedules. The latter is a formal continuum extrapolation in the native regime, where thin walls invalidate its spatial-gradient assumptions; it is not a validated native transport coefficient. Bottom right: independent $L=256$ validation of the late-window coarsening closure, either transferring both coefficients from $L=128$ or fitting on the designated $L=256$ training interval. Bars are $95\%$ trajectory-bootstrap intervals. This calibrated network closure is distinct from the no-fit controlled geometry tests.}
 \label{fig:wall-driven}
\end{figure}

We next drive six circular domains from $t=t_c-0.03$ to $t_c-0.3$, using radii $40,56$ and rates $\gamma=10^{-3},3\times10^{-3},10^{-2}$. The initial field is the exact homogeneous amplitude times a tanh profile of width $\sqrt{2\kappa/(-r)}$. Unlike the frozen-circle measurement, which integrates a diffuse phase indicator, this experiment locates the zero contour by linear interpolation along each lattice row. Its enclosed area defines $R_{\rm contour}^2=A/\pi$. The leading prediction overestimates the loss of squared radius by $13.4$--$15.7\%$; replacing $\kappa$ by the independently computed $\kappa_{\rm wall}$ reduces these errors to $0.10$--$2.13\%$ (Fig.~\ref{fig:wall-driven}). This is a forward test of the integrated displacement from the prepared initial field, including its transient, rather than a fit of an effective motion coefficient. No uniform adiabatic error bound is established for these finite ramps.

\subsubsection{Domain networks and native generation.}
For irregular domains, curvature motion motivates $L_{\rm dom}^2=b+C\int\kappa\,d\tau$, with a morphology-dependent $C$. We test this closure on $24$ independent weak-noise trajectories at $\gamma=10^{-3}$, $D=2\times10^{-5}$, $L=128$, from $t_c+0.2$ to $t_c-0.35$, and $64$ native trajectories at $\gamma=1$, $D=1/2$, $L=128$, from $t=8$ to $0.08$. Native initial fields are independent standard Gaussian pixels; slow initial fields use the stable exact linear stationary covariance. The terminal cutoff is finite: these are not simulations at $t=0$. We measure inverse nearest-neighbor sign-boundary bond density, with Gaussian smoothing widths $0,1,2$ sites, rather than equating this estimator with spectral curvature or connected-domain diameter. Specifically, for $h_i=\operatorname{sign}(x_i)$ after the specified smoothing, the estimator is $[p_x+p_y]^{-1}$, where $p_\mu$ is the fraction of nearest-neighbor bonds with opposite signs in direction $\mu$. It counts both lattice directions and is sensitive to wall orientation. The fitted closure uses the square of the ensemble mean length, not the ensemble mean of squared individual lengths.

Calibrating $b,C$ over $t_c-0.20<t<t_c-0.08$ and predicting later times gives squared-length errors of $21$--$46\%$ in the slow ensemble. For native generation, calibration over $1.5<t<3$ gives later errors of $22$--$200\%$, depending on smoothing. The apparent agreement of the raw estimator is insufficient evidence for a wall model because microscopic sign fluctuations dominate it early in the trajectory. The large estimator dependence and rapid drive prevent a quantitative validation of the sharp-wall reduction in native sampling.

\subsubsection{Separating saturation from coarsening.}
The earlier slow calibration interval includes substantial amplitude growth: at $t=t_c-0.1$, the ensemble RMS is only about $16\%$ of the exact homogeneous amplitude, whereas it reaches about $91\%$ at $t_c-0.2$. An exploratory calibration restricted to $t_c-0.28<t<t_c-0.20$ gives later squared-length residuals of $1.2$--$2.1\%$ with $\kappa$, and $0.75$--$1.47\%$ with $\kappa_{\rm wall}$. This selected interval cannot independently validate the closure. We therefore fix that protocol before generating a new $24$-trajectory ensemble at $L=256$, with the same rate, noise, timestep and time range but independent seed $20261020$. Fitting on its designated training interval gives $1.01$--$1.08\%$ error on its later observations, with $95\%$ whole-trajectory bootstrap intervals extending approximately from $0.2\%$ to $2.6\%$. Transferring both fitted coefficients directly from $L=128$, without refitting, instead gives $10.5$--$11.2\%$ error, with broad bootstrap intervals extending up to approximately $40\%$. The functional form is therefore useful after saturation, conditional on calibration; these data do not establish a system-size-independent coefficient. Sampling and finite-size effects cannot be separated by this single independent size comparison.

\subsubsection{Numerics and reproducibility.}
The fixed and driven deterministic geometries use a midpoint drift step of $0.1$. Refining a frozen circle at $t_c-t=0.03$, $R=40$, $L=128$, for duration $100$, from $0.1$ to $0.05$ changes its diffuse squared-radius estimate by less than $10^{-7}$ sites squared. This is a representative check, not convergence of every driven geometry. Stochastic drift also uses an explicit midpoint step, with additive Brownian increments and coefficients evaluated at the beginning of each step; schedule discretization remains first order. Slow ensembles use $\Delta\tau=0.15$ and native ensembles $0.005$. Eight slow trajectories on $64^2$ with Brownian-coupled steps $0.15/0.075$ give a terminal relative field RMS difference of $0.144\%$ and identical sign-length estimates. Sixteen native trajectories on $64^2$ with coupled steps $0.005/0.0025$ give a terminal relative RMS difference of $0.980\%$ and raw sign lengths $6.9383/6.9607$, a $0.32\%$ change. These checks do not certify every time-resolved wall observable or the larger ensembles. Bootstrap intervals resample complete trajectories ($300$ exploratory or $400$ independent-validation replicates). 

\clearpage

% END KZ SUPPLEMENT ADDITION

\section{Additional results}
\begin{figure}[htbp]
\centering
\includegraphics[width=\linewidth]{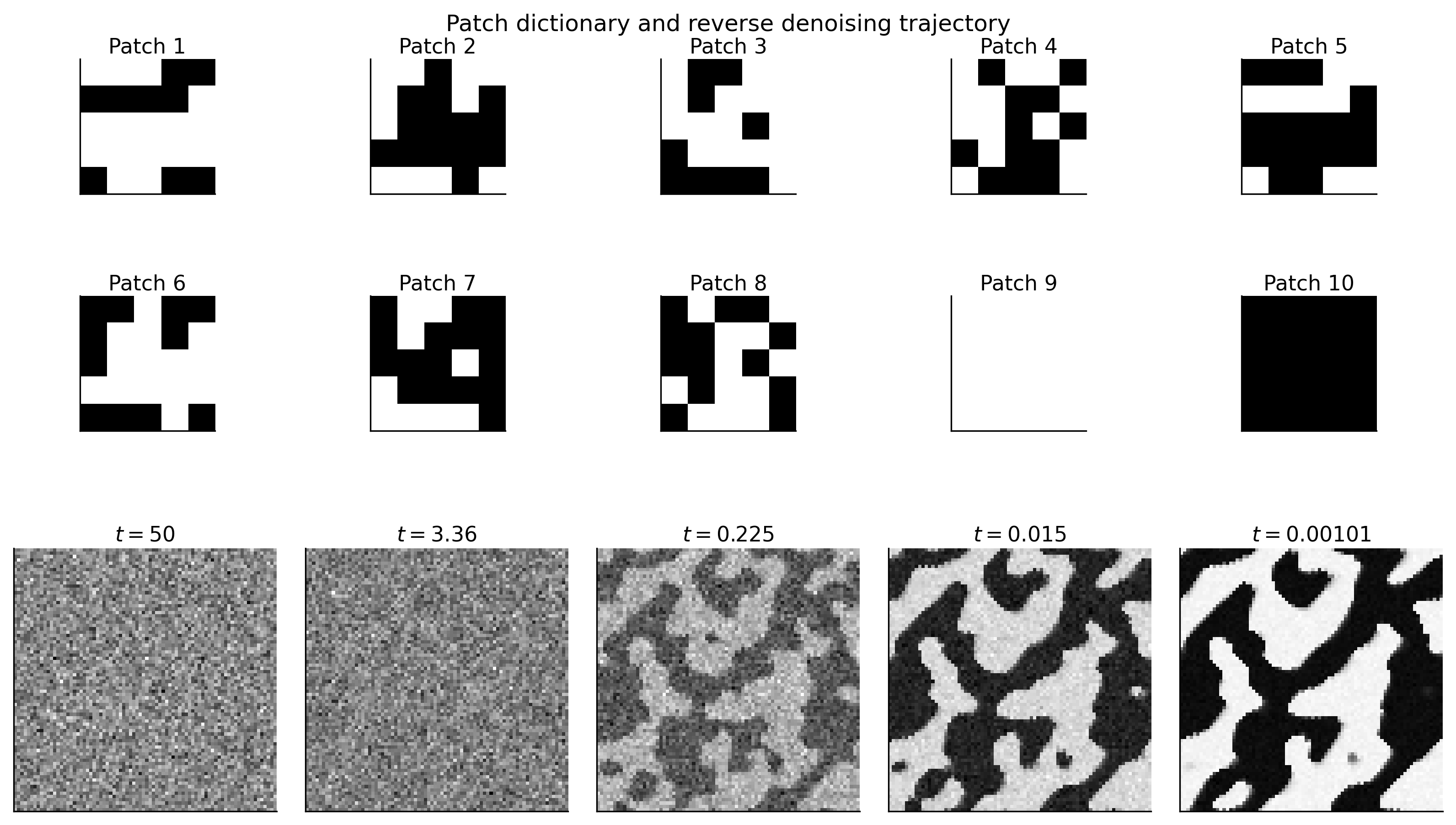}
\caption{
Patch dictionary (top) and a representative reverse denoising trajectory (bottom). 
Five snapshots of the evolving configuration are shown at logarithmically spaced times along the reverse diffusion path. 
Initially the field is dominated by noise, while at later times large coherent domains form as the dynamics selects one of the dominant global patterns.
}
\label{fig:patch_results_full}
\end{figure}

\begin{figure}[htbp]
\centering
\includegraphics[width=\linewidth]{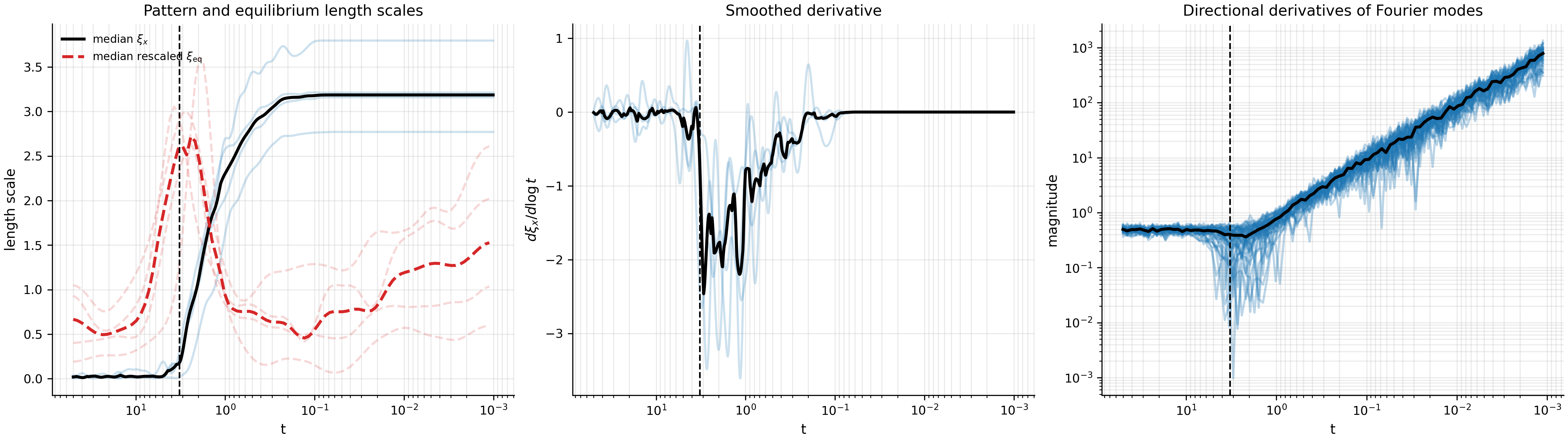}
\caption{
Pattern-correlation and Fourier-response diagnostics for the binarized Fashion-MNIST model. The methods settings are described in Appendix~\ref{supp: networks training}.
}
\label{fig:fmnist_results}
\end{figure}

\begin{figure}[htbp]
\centering
\includegraphics[width=\linewidth]{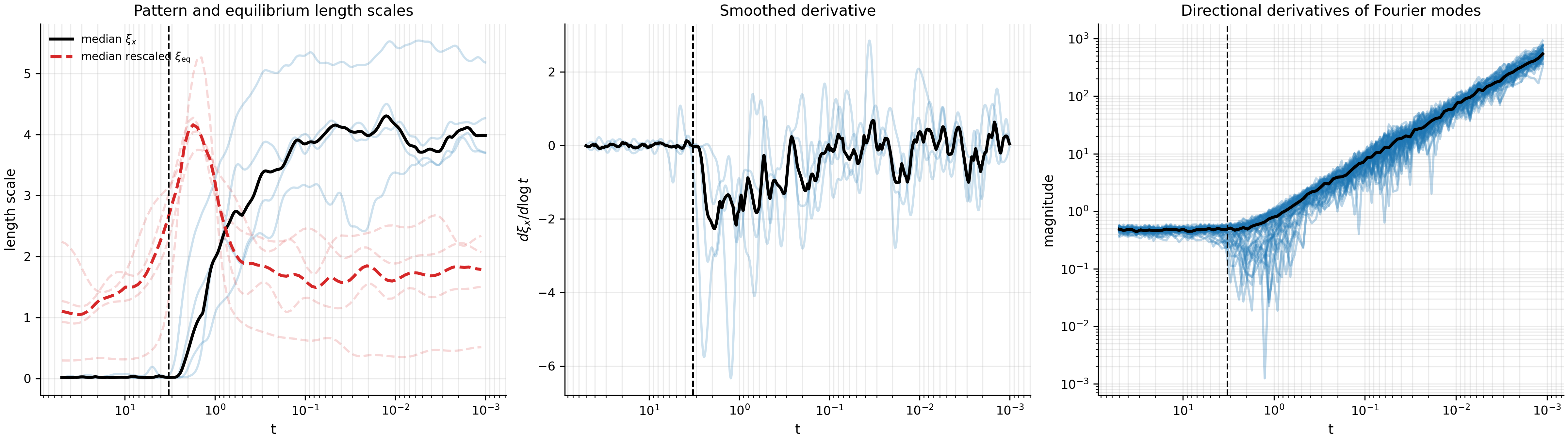}
\caption{
Quantitative analysis of pattern formation dynamics on non-binarized ConvNets trained on FashionMNIST. The critical line uses the binary-model reference and is only approximate for continuous-valued data. 
}
\label{fig:fmnist_soft_results}
\end{figure}

%\newpage
%\input{checklist.tex}

\end{document}